\documentclass[12pt]{l4dc2022}

\title[Safe Control with NNDM]{Safe Control with Neural Network Dynamic Models}

\usepackage{xcolor}
\usepackage{amsmath,amsfonts,amssymb,mathtools}

\newcommand{\FF}{\mathcal{F}}
\newcommand{\GG}{\mathcal{G}}

\newcommand{\XX}{\mathcal{X}}

\newcommand{\real}{\mathbb{R}}

\newcommand{\ie}{\textit{i.e. }}
\newcommand{\st}{\textit{s.t. }}

\renewcommand{\v}[1]{\boldsymbol{\mathbf{#1}}}
\newcommand{\dotv}[1]{\dot{\boldsymbol{\mathbf{#1}}}}

\newcommand{\cor}[1]{{\color{orange} #1}}
\newcommand{\cbl}[1]{{\color{blue} #1}}

\newcommand{\cgr}[1]{{\color{teal} #1}}

\usepackage{times}
\usepackage{cleveref}
\usepackage{comment}
\usepackage{booktabs}
\usepackage{multirow}
\usepackage[font = footnotesize]{caption}

\author{%
 \Name{Tianhao Wei} \Email{twei2@andrew.cmu.edu}\\
 \addr Carnegie Mellon University
 \AND
 \Name{Changliu Liu} \Email{cliu6@andrew.cmu.edu}\\
 \addr Carnegie Mellon University%
}

\begin{document}

\maketitle

\begin{abstract}%
Safety is critical in autonomous robotic systems. A safe control law should ensure forward invariance of a safe set (a subset in the state space). It has been extensively studied regarding how to derive a safe control law with a control-affine analytical dynamic model. 
However, how to formally derive a safe control law with Neural Network Dynamic Models (NNDM) remains unclear due to the lack of computationally tractable methods to deal with these black-box functions. 
In fact, even finding a control that minimizes an objective for NNDM without any safety constraint is still challenging. In this work, we propose MIND-SIS (Mixed Integer for Neural network Dynamic models with Safety Index Synthesis), the first method to synthesize safe control for NNDM. The method includes two parts: 1) SIS: an algorithm for the offline synthesis of the safety index (also called as a barrier function) using evolutionary methods and 2) MIND: an algorithm that computes the optimal safe control input online by solving a constrained optimization with a computationally efficient encoding of neural networks. It has been theoretically proved that MIND-SIS guarantees forward invariance and finite-time convergence to a subset of the user-defined safe set. It has also been numerically validated that MIND-SIS achieves optimal safe control of NNDM with less than $10^{-8}$ optimality gap and zero safety constraint violation.
\end{abstract}


\section{Introduction}

Robot safety depends on the correct functioning of all system components, such as accurate perception, safe motion planning, and safe control. Safe control, as the last defense of system safety, has been widely studied in the context of dynamical systems~\cite{nagumo1942lage, blanchini1999set}. 
A safe control law ensures the forward invariance of a subset inside the user-defined safety constraint, meaning that any agent entering that subset will remain in it. There are many methods to derive the safe control laws for control-affine analytical dynamic model~\cite{wei2019safe, liu2014control}. However, constructing such an analytical dynamic model for complex systems can be difficult, time-consuming, and sometimes impossible~\cite{nguyen2011model}. Recent works adopt data-driven approaches to learn these dynamic models, and most of the learned models are encoded in neural networks, e.g. virtual world models of video games or dynamic models of a robot, etc. ~\cite{nagabandi2018neural, janner2019trust}. 
Although neural network dynamic models (NNDMs) can greatly alleviate human efforts in modeling, they are less interpretable than analytical models. It is more challenging to derive control laws, especially safe control laws, for these NNDMs than for analytical models.

This paper focuses on safe tracking tasks with NNDMs, which is formulated as a constrained optimization that minimizes the state tracking error given the safety constraint and the neural network dynamics constraint. 
Even without the safety constraint, the tracking control with NNDMs is already challenging. 
Since NNDMs are complex and highly nonlinear, there is no computationally efficient method to compute its model inverse, which is required by most existing white-box methods~\cite{tolani2000real}. 
On the other hand, black-box methods, such as the shooting method which chooses control from randomly generated candidates, can not guarantee to find the optimal solution in finite time.
Moreover, the safety constraint adds another layer of difficulty to the problem. 
The robot should select an action that not only satisfies the safety constraint at the current time step, but also ensures that in the future, the agent will not enter any state where no action is safe. This property is called \textit{persistent feasibility}. To ensure persistent feasibility, we need to compute the control invariant set inside the original user-specified safety constraint and constrain the robot motion in this more restrictive control invariant set. 
For an analytical model, we can manually craft this control invariant set to meet the requirement~\cite{liu2014control} based on our understanding of the dynamics. The same task becomes difficult for NNDM due to its poor interpretability.

In this work, we address these challenges by introducing an integrated method, mixed integer for neural network dynamic models with safety index synthesis (MIND-SIS), to handle both the offline synthesis of the control invariant set and the online computation of the constrained optimization with NNDM constraints.
First, inspired by an algorithm for neural network verification~\cite{tjeng2017evaluating}, we use mixed integer programming (MIP) to encode the NNDM constraint, which greatly reduces the complexity of the optimization problem. Importantly, the MIP method is complete and guarantees optimality.
Second, to synthesize the control invariant set, we use evolutionary algorithms to optimize a parameterized safety index.
Using the learned safety index, the resulting control solved by the constrained optimization will ensure persistent feasibility and hence forward invariance inside the user-specified safety constraint. 


The remaining of the paper is organized as follows. \Cref{sec:formulation} provides a formal description of the problem and introduces notations. \Cref{sec:related_work} introduces prior works on safe control, NNDM, and neural network verification, which inspire our method. \Cref{sec:method} discusses the proposed method in detail. \Cref{sec:exp} shows experimental results that validate our method. And \cref{sec:discussion} discusses possible future directions. Additional results and discussions can be found in the appendix in the arxiv version \url{https://arxiv.org/abs/2110.01110}. The code is at \url{https://github.com/intelligent-control-lab/NNDM-safe-control}.










\section{Formulation} \label{sec:formulation}


\paragraph*{Dynamic model}
        
        Consider a discrete time dynamic system with $m_x$ state and $m_u$ controls.
        \begin{align}
            \quad \v x_{k+1} = \v x_k + \v f(\v x_k, \v u_k) dt,
        \end{align}
        where $k$ is the time step, $\v x_k \in X \subset \real^{m_x}$ is the state, $\v u_k \in U \subset \real^{m_u}$ is the control, $\v f: \real^{m_x} \mapsto \real^{m_u}$ is the dynamic model, and $dt$ is the sampling time. We assume the legal state set $X$ and control set $U$ are both defined by linear constraints. This assumption covers most cases in practice.
        
        In the NNDM case, the dynamic model $\v f$ is encoded by a $n$-layer feedforward neural network. Each layer in $\v{f}$ corresponds to a function $\v f_i:\real^{k_{i-1}}\mapsto \real^{k_i}$, where $k_i$ is the dimension of the hidden variable $\v{z}_i$ in layer $i$, and $k_0 = m_x+m_u$, $k_n = m_x$. The network can be represented by $\v{f} = \v{f}_n \circ \v{f}_{n-1} \circ \cdots \circ \v{f}_1$,
        where $\v{f}_i$ is the mapping for layer $i$. And $\v z_i = \v{f}_i(\v z_{i-1}) = \v{\sigma}_i(\hat{\v z}_i) = \v{\sigma}_i(\v W_i \v z_{i-1} + \v b_i)$ 
        where $\v W_i \in \real^{k_i\times k_{i-1}}$ is the weight matrix, $\v b_i\in \real^{k_i}$ is the bias vector, and $\v \sigma_i: \real^{k_{i}}\mapsto \real^{k_i}$ is the activation function. We only consider ReLU activation in this work. For simplicity, denote $\v W_i \v z_{i-1} + \v b_i$ by $\hat{\v z}_i$.
        Let $z_{i,j}$ be the value of the $j^{th}$ node in the $i^{th}$ layer, $\v w_{i,j} \in \real^{1\times k_{i-1}}$ be the $j^{th}$ row in $\v W_i$, and $b_{i,j}$ be the $j^{th}$ entry in $\v b_j$. 
        
\paragraph*{Safety specification}
We consider the safety specification as a requirement that the system state should be constrained in a connected and closed set $\XX_0 \subseteq X$. $\XX_0$ is called the safe set. $\XX_0$ should be a zero-sublevel set of an initial safety index $\phi_0: X \mapsto \real$, \ie $\XX_0 = \{\v x \mid \phi_0(\v x) \leq 0\}.$
$\phi_0$ can be defined differently for a given $\XX_0$. Ideally, a safe control law should guarantee forward invariance and finite-time convergence to the safe set. Forward invariance requires that $\phi_0(\v x_{k}) \leq 0 \implies \phi_0(\v x_{k+1}) \leq 0$.
And finite-time convergence can be enforced by requiring that $\phi_0(\v x_k) > 0 \implies \phi_0(\v x_{k+1}) \leq \phi_0(\v x_{k}) - \gamma dt$. Hence the number of time steps for an unsafe state to return to the safe set is bounded above by $\phi_0(\v x_k)/\gamma dt$.
These two conditions can be written compactly as one:
\begin{align}
    \phi_0(\v x_{k+1}) \leq \max\{0, \phi_0(\v x_{k}) - \gamma dt\}.\label{eq:safety}
\end{align}

\paragraph*{The safe tracking problem}
This paper considers the following constrained optimization for safe tracking, where the problem is solved at every time step $k$: 
\begin{align}
    \label{eq:safe_control}
    \begin{split} 
        \min_{\v u_k,\v x_{k+1}} & \|\v x_{k+1} - \v x_{k+1}^r\|_p\\ 
        \st & \v x_{k+1} = \v x_k + \v f(\v x_k, \v u_k) dt,\quad \v u_k\in U\\
         & \phi_0(\v x_{k+1}) \leq \max\{0, \phi_0(\v x_{k}) - \gamma dt\}\\
    \end{split}
\end{align}
where $\v x_{k+1}^r$ is the reference state at time step $k+1$, $\|\cdot\|_p$ can be either $\ell1$-norm or $\ell2$-norm. This formulation can be viewed as a one-step model predictive control (MPC). The extension to multi-step MPC is straightforward, which we leave for future work. 
At a given step $k$, \eqref{eq:safe_control} is a nonlinear programming problem. However, existing nonlinear solvers have poor performance for constraints involving neural networks (which will be shown in \cref{sec:exp}). The reason is that neural networks (with ReLU activation) are piece-wise linear, whose second-order derivatives are not informative. New techniques are needed to solve this problem.

\paragraph{Persistent feasibility}
Persistent feasibility requires that there always exists $\v u_k\in U$ that satisfies \eqref{eq:safety} for all time step $k$. However, this may not be true for some $\v x_k \in \XX_0$.
For example, if $\phi_0$ measures the distance between the ego vehicle and the leading vehicle. It is possible that the ego vehicle is still far from the leading vehicle ($\phi_0(\v x_k) < 0$), but has big relative speed toward the leading vehicle. Then collision is inevitable ($\phi_0(\v x_{k+1}) > 0$ for all possible $\v u_k\in U$).
This situation may happen when the relative degree from $\phi_0$ to $\v u$ is greater than one or when the control inputs are bounded. 
In these cases, $\XX_0$ may not be forward invariant or finite-time convergent. We call this situation as \textit{loosing control feasibility}, which further leads to \textit{loosing persistent feasibility}. 
To address this problem, we want to prevent the system from getting into those control-infeasible states in $\XX_0$. That is to find a subset $\XX_s \subseteq\XX_0$ such that there exists a feasible control law to make $\XX_s$ forward invariant and finite-time convergent. We call $\XX_s$ a \textit{control invariant set} within $\XX_0$. 



\section{Related work} \label{sec:related_work}
\paragraph*{Optimization with Neural Network Constraints}
Recent progress in nonlinear optimization involving neural network constraints can be classified as primal optimization methods and dual optimization methods. 
The primal optimization methods encode the nonlinear activation functions (e.g., ReLU) as mixed-integer linear programmings \cite{tjeng2017evaluating}, relaxed linear programmings \cite{ehlers2017formal} or semidefinite programmings \cite{raghunathan2018semidefinite}. 
Our method to encode NNDM is inspired by MIPVerify \cite{tjeng2017evaluating}, which uses mixed integer programming to compute maximum allowable disturbances to the input. MIPVerify is complete and sound, meaning that the encoding is equivalent to the original problem. 


\paragraph*{QP-based safe control}
When the system dynamics are analytical and control-affine, the safe tracking problem can be decomposed into two steps: 1) computing a reference control $\v u^{r}$ without the safety constraint; 2) projecting $\v u^r$ to the safe control set~\cite{ames2019control}. 
For analytical control-affine dynamic models, the safe control set that satisfies \eqref{eq:safe_control} is a half-space intersecting with $U$. Therefore, 
the second step is essentially a quadratic projection of the reference control to that linear space, which can be efficiently computed by calling a quadratic programming (QP) solver. Existing methods include CBF-QP~\cite{ames2016control}, SSA-QP~\cite{liu2014control}, etc.
However, to our best knowledge, there has not been any quadratic projection method that projects a reference control to a safety constraint with non-analytical and non-control-affine dynamic models, in which case the safe control set can be non-convex. Moreover, our work solves both the computation of reference control and its projection onto safe control set in an integrated manner. Hence, our work is not limited to the quadratic projection of the reference control.

\paragraph*{Persistent feasibility in MPC}
There are different approaches in MPC literature to compute the control invariant set to ensure persistent feasibility, such as Lyapunov function~\cite{danielson2016path}, linearization-convexification~\cite{jalalmaab2017guaranteeing}, and grid-based reachability analysis~\cite{bansal2017hamilton}. However, most of the non-grid-based methods approximate the control invariant set by convex set, which greatly limit the expressiveness of the geometries. Although grid-based methods have better expressiveness and may be able to extend to non-analytical models, they have limited scalability due to the curse of dimensionality and the fact they are usually non-parameterized. Our method can synthesize the control invariant set with nonlinear boundaries for non-analytical models using parameterized functions, hence more computationally efficient. 

\section{Method} \label{sec:method}

In this section, we discuss how to efficiently solve the constrained optimization \eqref{eq:safe_control} and ensure it is persistently feasible. First, we introduce MIND, a way to find the optimal solution of \cref{eq:safe_control} by encoding NNDM constraints as mixed integer constraints. Then we present SIS, a method to find the control invariant set by learning a new safety index $\phi$ that maximizes control feasibility. Finally, we present the reformulated problem.

\subsection{MIND: Encode NNDM constraints}
To overcome the complexity of NNDM constraints, we first add all hidden nodes in the neural network as decision variables and turn \eqref{eq:safe_control} into the following equivalent form:
\begin{align}\label{eq: safe control equivalent form}
    \begin{split} 
        \min_{\v u_k,\v x_{k+1},\v z_i} & \|\v x_{k+1} - \v x_{k+1}^r\|_p\\ 
        \st & \v x_{k+1} = \v x_k + \v z_n dt, \v z_0 = \left[\v x_k, \v u_k\right], \quad \v u_k\in U\\
        & z_{i, j}=\max\{\hat z_{i,j}, 0\}, \hat z_{i,j} = \mathbf{w}_{i, j} \mathbf{z}_{i-1}+b_{i, j},\forall i \in\{1, \ldots, n\}, \forall j \in\left\{1, \ldots, k_{i}\right\}\\
         & \phi_0(\v x_{k+1}) \leq \max\{0, \phi_0(\v x_{k}) - \gamma dt\}.
    \end{split}
\end{align}

Nevertheless, the nonlinear non-smooth constraints introduced by the ReLU activation $z_{i, j}=\max\{\hat z_{i,j}, 0\}$ in \eqref{eq: safe control equivalent form} is still challenging to handle. 
Inspired by MIPVerify~\cite{tjeng2017evaluating}, we use mixed integer formulation to rewrite these constraints. We first introduce an auxiliary variable $\delta_{i,j}$ to denote the activation status of the ReLU node:
\begin{align}
    \delta_{i, j}=1 \Rightarrow z_{i, j}=\hat{z}_{i, j}, \quad
    \delta_{i, j}=0 \Rightarrow z_{i, j}=0.
\end{align}
Then we compute the pre-activation upper bounds $\hat u_{i,j}$ and lower bound $\hat l_{i,j}$ of every node in the neural network using interval arithmetics~\cite{moore2009introduction}. Given the input ranges (e.g., $x \in [1,2]$ and $y \in [3,4]$),
interval arithmetics compute the output range using the lower and upper bounds (e.g., $x-y \in [1-4,2-3] = [-3,-1]$). When $\hat{u}_{i, j} \leq 0$, the constraint for ReLU activation reduces to $z_{i,j} = 0$. When $\hat{l}_{i,j} \geq 0$, the constraint reduces to $z_{i,j} = \hat{z}_{i,j}$. Otherwise, the constraint can be represented as the following linear inequalities~\cite{liu2020algorithms}: 
\begin{align}
    \label{eq:network_encoding}
    z_{i, j} \geq \hat{z}_{i, j}, 
    z_{i, j} \geq 0, 
    z_{i, j} \leq \hat{z}_{i, j}-\hat{\ell}_{i, j}\left(1-\delta_{i, j}\right), 
    z_{i, j} \leq \hat{u}_{i, j} \delta_{i, j},
    \delta_{i,j} \in \{0,1\}.
\end{align}

\begin{figure}[t]
     \centering
     \subfigure{\label{main:a}\includegraphics[width=.32\linewidth]{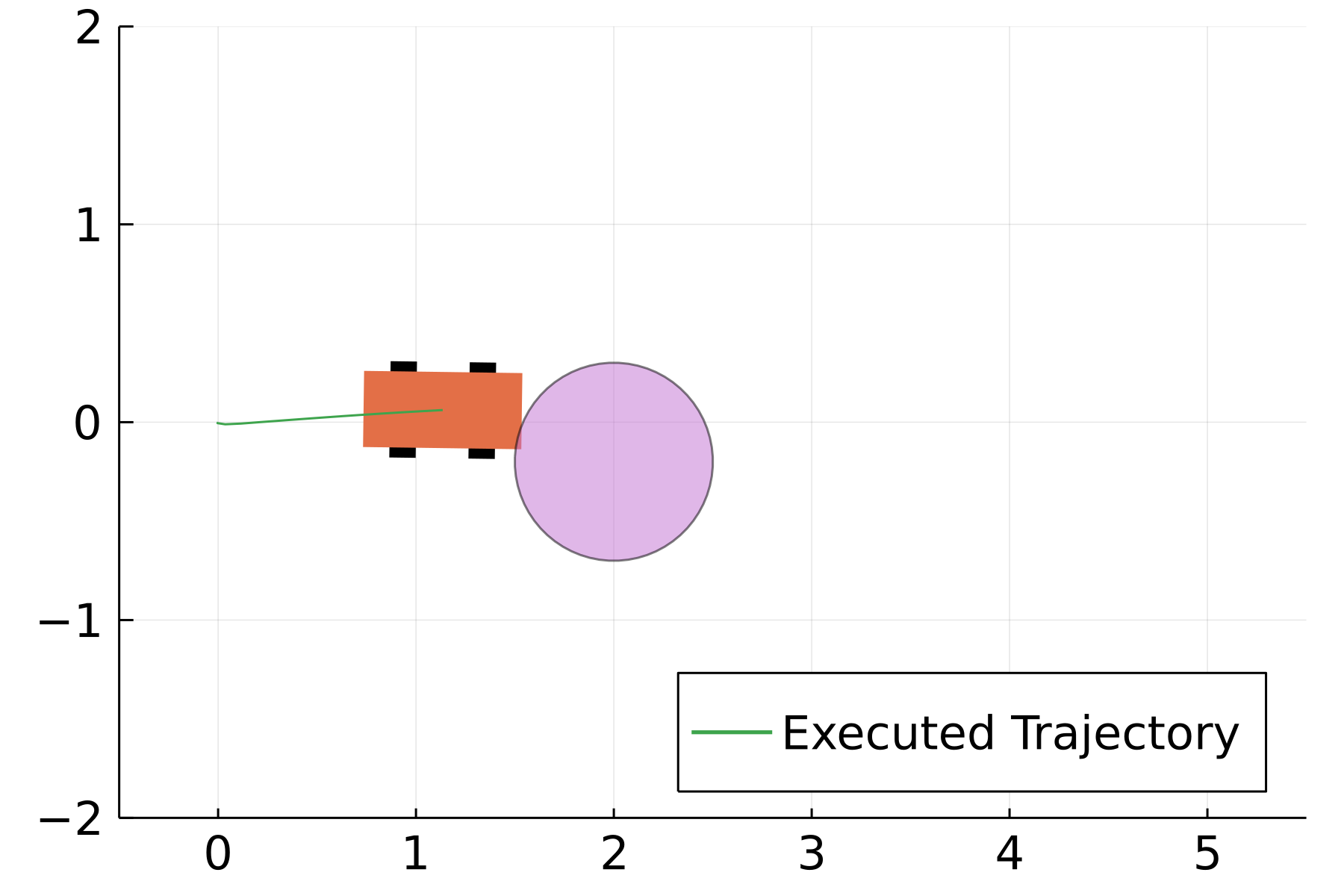}}
     \hfill
     \subfigure{\label{main:b}\includegraphics[width=.32\linewidth]{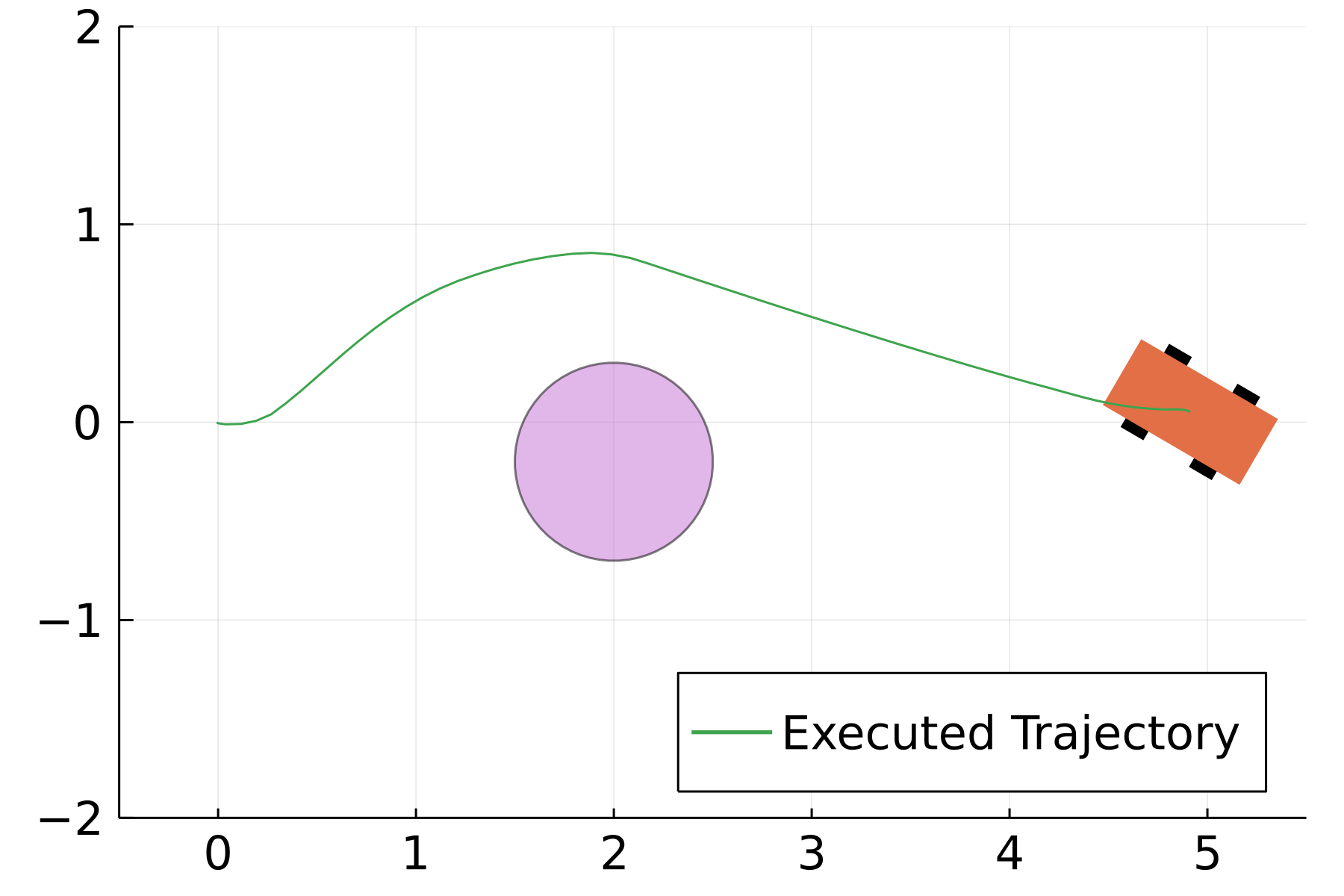}}
     \hfill
     \subfigure{\label{main:c}\includegraphics[width=.32\linewidth]{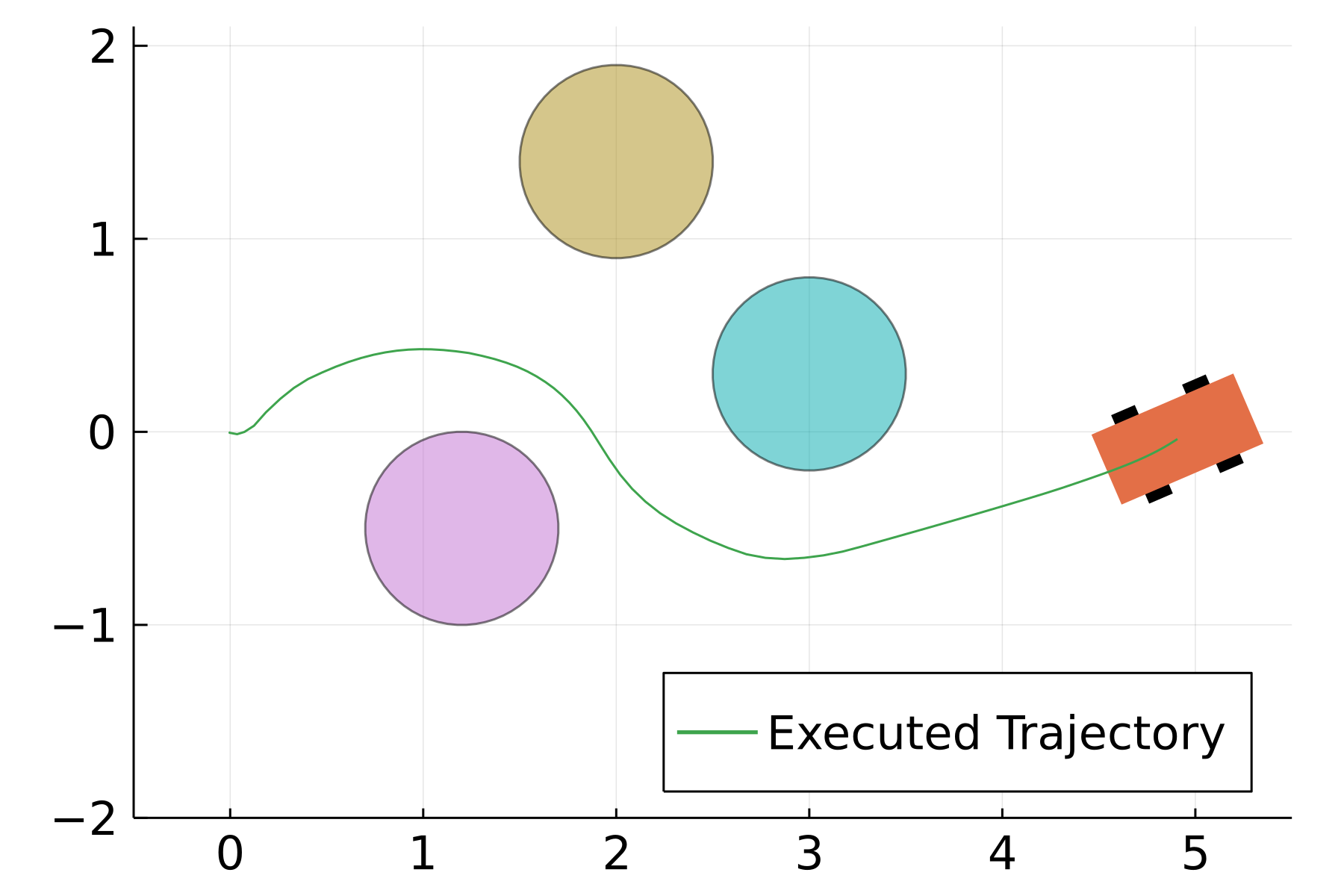}}
     \caption{MIND-SIS in collision avoidance. (a) shows the performance of MIND with $\phi_0$. The vehicle collides with the obstacle due to loss of control feasibility (too late to break). (b) and (c) show the performance of MIND-SIS. The SIS-synthesized safety index $\phi^l$ guarantees persistent feasibility and is general enough to be directly applied to multi-obstacle scenarios  without any modification.}
     \vspace{-5mm}
     \label{fig:obstacle}
\end{figure}

With this encoding, the constrained optimization is converted into a MIP, which can be solved efficiently. Theoretically, MIP is a NP-complete problem. The worst case computation time grows exponentially with the number of integer variables, which is the total number of ReLU activation functions. However, in practice, the computation can be greatly accelerated by various techniques developed in recent years \cite{gurobi}. The evaluation in \cref{sec:exp} shows that the actual computation time for a network with $100$ ReLUs is only $0.36$ seconds while that for a network with $200$ ReLUs is $0.8$ seconds. Besides, it is worth noting that MIND scales well with dimensions of state and control when the total number of neurons in the hidden layers is fixed.

Nevertheless, successfully obtaining the solution for time $k$ and executing the control does not necessarily ensure we will have a solution to the constrained optimization in future time steps as shown in \cref{fig:obstacle}(a). The safety constraint needs to be modified to ensure persistent feasibility.

\subsection{SIS: Guaranteed persistent feasibility}

Unlike analytical models, it is challenging to design a safety index for NNDM because of its poor interpretability. Therefore, we introduce Safety Index Synthesis (SIS), which automatically synthesizes a safety index $\phi$ that results in a forward invariant and finite-time convergent $\XX_s$ to guarantee persistent feasibility.

\cite{liu2014control} introduced a form of $\phi$ that can improve control feasibility, $\phi(\v x) = \phi_0^*(\v \alpha_0, \v x) + \sum_{i=1}^q \alpha_i \phi_0^{(i)}(\v x) + \beta$, where $\phi_0^*(\v \alpha_0, \v x)$ defines the same sublevel set as $\phi_0$ and is parameterized by $\v \alpha_0$, $\phi_0^{(i)}(\v x)$ is the $i$-th order derivative of $\phi_0$, $q$ is the order such that the relative degree from $\phi_0^{(q)}$ to $\v u$ is $1$, and $\beta$ is a constant. We denote the concatenation $[\v \alpha_0, \alpha_1, \cdots, \alpha_q]$ by $\v \alpha$. 
\cite{liu2014control} showed that this $\phi$ could result in a forward invariant and finite-time convergent $\mathcal{X}_s$ when the control input is unbounded. When the control input is bounded, we argue that persistent feasibility can be achieved  by optimizing $\v \alpha$ and $\beta$ under \assumptionref{asp: lipschitz}. 

\begin{assumption}\label{asp: lipschitz}
$\v f$ and $\phi$ are Lipschitz continuous functions with Lipschitz constants  $k_f$ and $k_\phi$ respectively. The Euclidean norm of $\v f(\v x, \v u)$ is bounded by $M_f$.
\end{assumption}

To ensure forward invariance, it suffices to enforce that there always exists a feasible control for all states near the zero-level set of $\phi$ (as proved in \cref{apd:Xs}). 
Define the state-of-interest set $B = \{\v x \mid |\phi(\v x)| \leq k_{\phi} M_{f} dt\}$ (states near the boundary $\phi = 0$) and infeasible-state-of-interest set $B^* = \{\v x \mid \v x\in B, \forall \v u, \phi(\v x + \v f(\v x, \v u) dt) > \max\{0, \phi(\v x ) - \gamma dt\} \}$. $B$ contains all the states that can cross the boundary $\phi=0$ in one step because $\|\v x_{k+1} - \v x_k\| \leq \|\v f(\v x, \v u)\| dt \leq M_f dt$. To achieve forward invariance, we need all states in $B$ to have feasible control (i.e., when $B^*$ is empty). 
Then the problem can be formulated as $\min_{\v \alpha, \beta} |B^*|/|B|$. The expression of the corresponding control invariant set $\XX_s$ is derived in \cref{apd:Xs}. We can also let $B=X$ to learn a safety index that further achieves finite time convergence of $\XX_s$ at the cost of a potentially more conservative policy (see \cref{apd:Xs}). Since the gradient from $|B^*|/|B|$ to $\v \alpha,\beta$ is usually difficult to compute, we use a derivative-free evolutionary approach, CMA-ES \cite{hansen2016cma} to optimize the parameters. CMA-ES runs for multiple generations. In each generation, the algorithm samples many parameter candidates (called members) from a multivariate Gaussian distribution and evaluates their performance. A proportion of candidates with the best performance will be used to update the mean and covariance of the Gaussian distribution. To evaluate the parameter candidates, we sample a subset $S \subset B$ and minimize the infeasible rate $r := \frac{|S\cap B^*|}{|S|}$ as a surrogate for the original objective function $|B^*|/|B|$. We prove that if the sampling is dense enough and $r=0$, the safety constraint with the learned safety index is guaranteed to be feasible for arbitrary states in $B$.

\begin{lemma}\label{lemma: feasibility}
Suppose 1) we sample a state subset $S \subset B$ such that  $\forall \v x \in B$, $\min_{\v x' \in S } \|\v x - \v x'\| \leq \delta$, where $\delta$ is an arbitrary constant representing the sampling density.; and 2) $\forall \v x' \in S$, there exists a safe control $\v u$, \st $\phi(\v x' + \v f(\v x',\v u)dt) \leq \max\{-\epsilon, \phi(\v x') - \gamma dt -\epsilon\}$,
where $\epsilon = k_{\phi} (1 + k_f dt) \delta$. Then $\forall \v x \in B, \exists \v u,\ \st$
\begin{align}\label{eq: lemma 1 inequality}
    \phi(\v x + \v f(\v x,\v u)dt) \leq \max\{0, \phi(\v x ) - \gamma dt\}.
\end{align}
\end{lemma}

\begin{proof}
According to condition 1), $\forall \v x\in B$, we can find $\v x'\in S$ such that $\|\v x-\v x'\|\leq \delta$. According to condition 2), for this $\v x'$, we can find $\v u$ such that $\phi(\v x' + \v f(\v x',\v u)dt) \leq \max\{0,\phi(\v x')-\gamma dt\}-\epsilon$. Next we show $\v x$ and $\v u$ satisfy \eqref{eq: lemma 1 inequality} using \cor{Lipschitz condition} and \underline{triangle inequality}.
\begin{subequations}
\begin{align}
     \phi(\v x + \v f(\v x,\v u)dt) & = \cor{\phi(\v x + \v f(\v x,\v u)dt) - \phi(\v x' + \v f(\v x',\v u)dt)} \cgr{ +\phi(\v x' + \v f(\v x',\v u)dt)}\\
    &\leq \underline{\cor{k_{\phi}\|\v x - \v x' + [\v f(\v x,\v u) - \v f(\v x',\v u)]dt\|}} \cgr{+\max\{0,\phi(\v x')-\gamma dt\}- \epsilon} \\
    & \leq \underline{k_{\phi}\cbl{\|\v x - \v x'\|} + k_{\phi}\cor{\|\v f(\v x,\v u) - \v f(\v x',\v u)\|}dt} + \max\{0,\phi(\v x')-\gamma dt\}- \epsilon\\
    &\leq k_{\phi} \cbl{\delta} + k_{\phi} \cor{k_f \delta} dt - \epsilon + \max\{0,\phi(\v x')-\gamma dt\} \\
    &= \max\{0,\phi(\v x')-\gamma dt\}. 
\end{align}
\end{subequations}
Hence \eqref{eq: lemma 1 inequality} is verified.
\end{proof}

The computation time of SIS depends on the number of CMA-ES iterations and the time spent in finding $S$ in each iteration. Although a rigorous proof is missing, the convergence rate of CMA-ES is empirically exponential~\cite{hansen2001completely}. We can find $S$ in each iteration by uniformly sampling $X$. This process can be time consuming if the state dimension is high (e.g. $n > 10$), but we may accelerate the process by high dimensional Breadth-First-Search, which we leave for future work.

\subsection{MIND-SIS: Safe control with NNDM}
Once the safety index is synthesized, we substitute $\phi_0$ with $\phi$ in \eqref{eq: safe control equivalent form} to guarantee persistent feasibility. To address the nonlinearity in the safety constraint \eqref{eq:safety}, we approximate it with first order Taylor expansion at the current state $\v x_k$  (\cref{apd:continuous_vs_discrete} discusses how the safety guarantee is preserved):
\begin{align}
    \phi(\v x_{k+1}) = \phi(\v x_{k}) + \nabla_{\v x}\phi \cdot \v f(\v x_k, \v u_k) dt + o(\| \v f(\v x_k, \v u_k) dt\|),\label{eq:taylor}
\end{align}
where $\lim_{dt\to 0} o(\|\v f(\v x_k, \v u_k) dt\|) = 0$.
Then \eqref{eq: safe control equivalent form} is transformed into a mixed integer problem:
\begin{align}\label{eq: mixed integer program}
    \begin{split} 
        \min_{\v u_k,\v x_{k+1},\v z_i,\delta_{i,j} \in \{0,1\}} & \|\v x_{k+1} - \v x_{k+1}^r\|_p\\ 
        \st & \v x_{k+1} = \v x_k + \v z_n dt, \v z_0 = \left[\v x_k, \v u_k\right], \quad \v u_k\in U,\\
        & z_{i, j} \geq \hat{z}_{i, j}, 
    z_{i, j} \geq 0, 
    z_{i, j} \leq \hat{z}_{i, j}-\hat{\ell}_{i, j}\left(1-\delta_{i, j}\right), 
    z_{i, j} \leq \hat{u}_{i, j} \delta_{i, j},
    \\
        &  \hat z_{i,j} = \mathbf{w}_{i, j} \mathbf{z}_{i-1}+b_{i, j},\forall i \in\{1, \ldots, n\}, \forall j \in\left\{1, \ldots, k_{i}\right\}\\
         & \nabla_{\v x}\phi \cdot \v f(\v x_k, \v u_k) \leq  \max\{-\dfrac{\phi(\v x_k)}{dt},-\gamma\}.\\
    \end{split}
\end{align}
Depending on the norm $\|\cdot\|_p$, \eqref{eq: mixed integer program} is either a Mixed Integer Linear Programming or Quadratic Programming, which both can be solved by existing solvers, such as GLPK, CPLEX, and Gurobi.

\begin{figure}[t]
     \centering
     \subfigure{\label{track:a}\includegraphics[width=45mm]{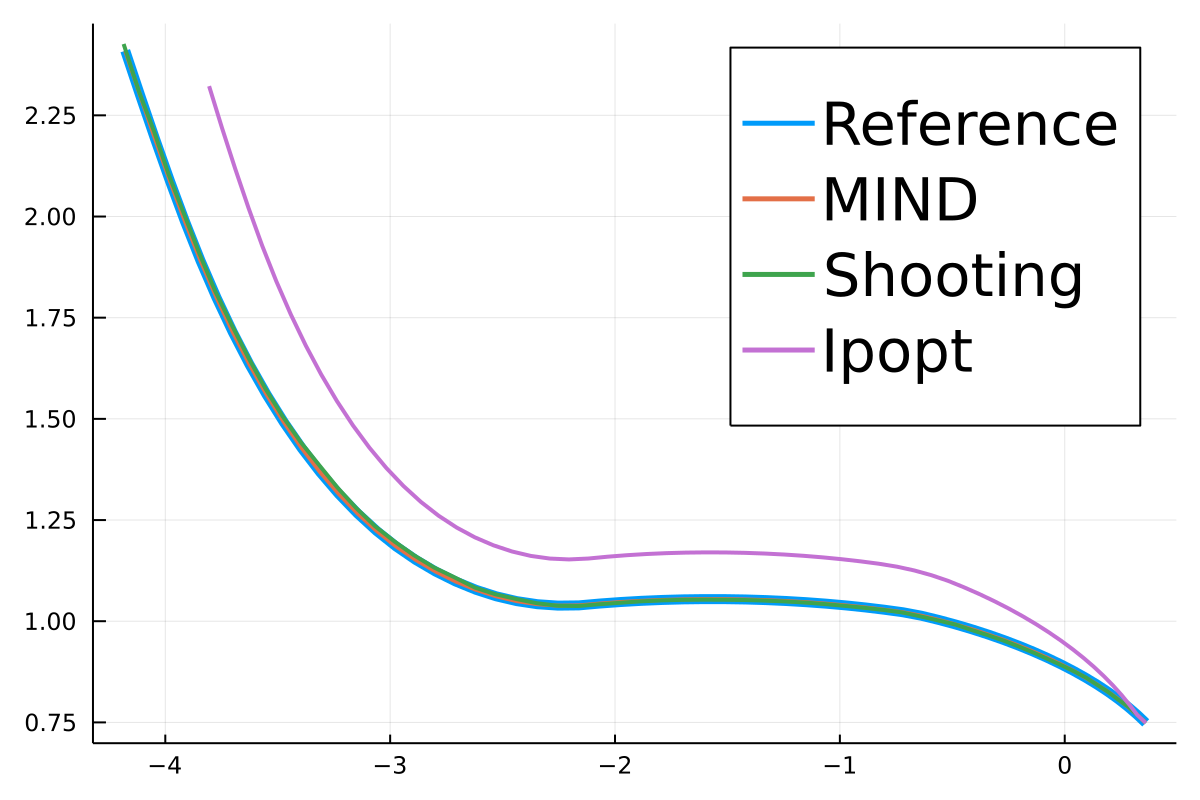}}
     \hfill
     \subfigure{\label{track:b}\includegraphics[width=45mm]{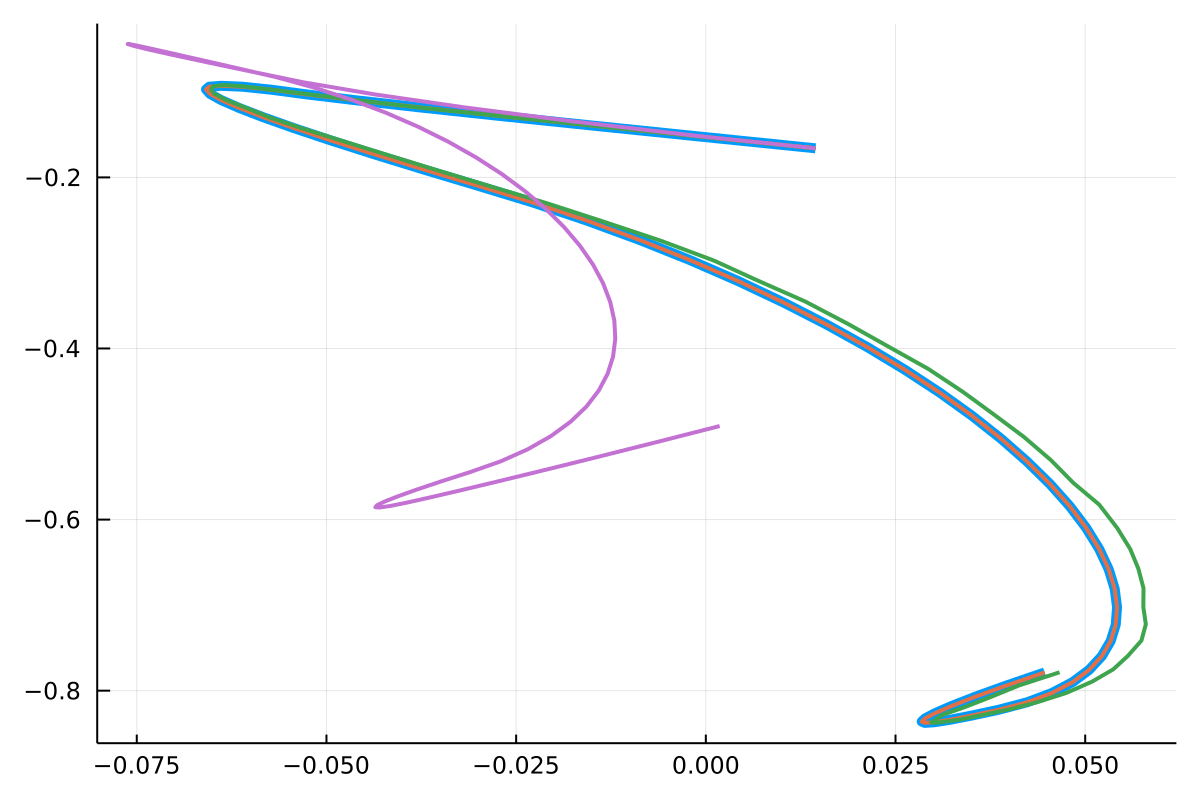}}
     \hfill
     \subfigure{\label{track:c}\includegraphics[width=45mm]{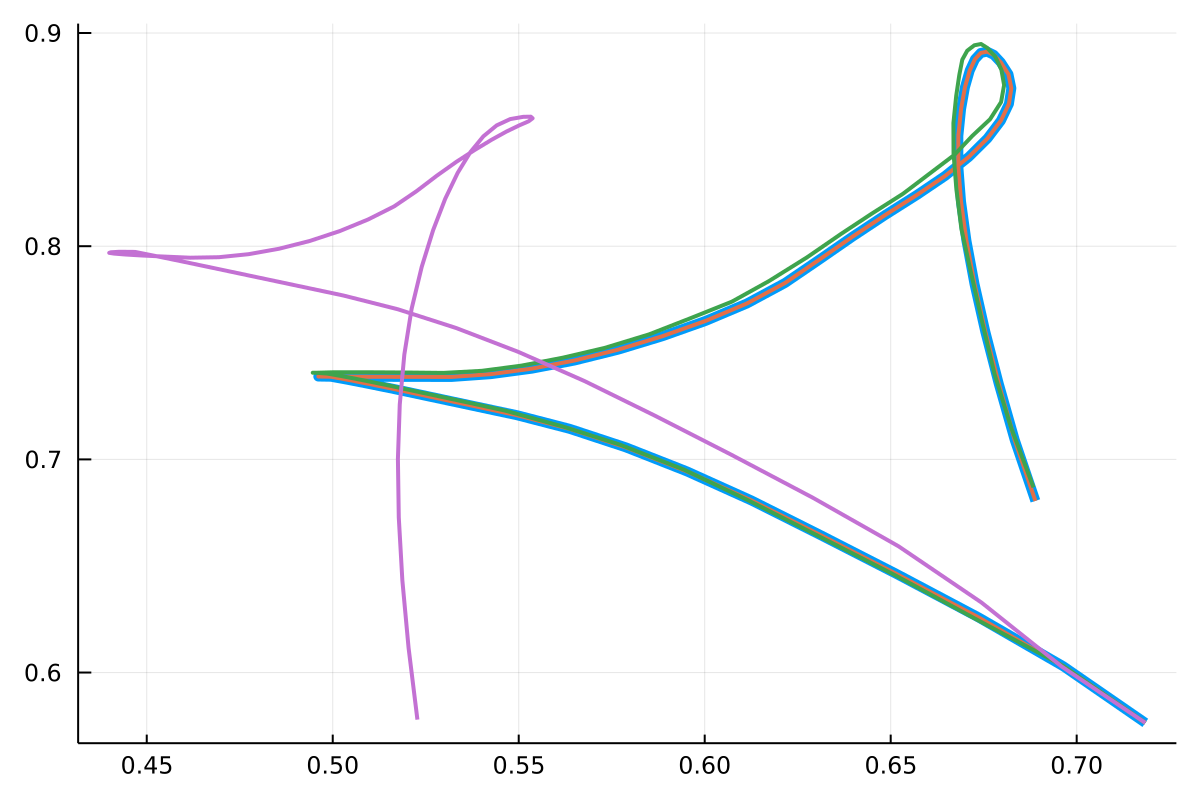}}
     \caption{Trajectory tracking with NNDM II using different optimization methods: MIND (our method), shooting method with sample size of 100, and Ipopt. (a-c) show 3 randomly generated trajectories. MIND has the smallest tracking error in all three cases.}
     \vspace{-5mm}
    \label{fig:tracking}
\end{figure}


\section{Experiment} \label{sec:exp}

\subsection{Experiment set-up}
The evaluation is designed to answer the following questions: 
1) How does our method (by solving \eqref{eq: mixed integer program}) compare to the shooting method and regular nonlinear solvers in terms of optimality and computational efficiency on problems without safety constraints?
2) Does the safety index synthesis improve persistent feasibility?
3) Does our method ensure safety in terms of forward invariance.

We evaluate our method on a system with  NNDMs for 2D vehicles. The NNDMs are learned from a second order unicycle dynamic model with 4 state inputs (2D position, velocity, and heading angle), 2 control inputs (angular velocity, acceleration), and 4 state outputs (2D velocity, angular velocity, and acceleration). All the states and controls are bounded, where $X: [-10,10]\times[-10,10]\times[-2,2]\times[-\pi,\pi]$ and $U: [-4,4]\times[-\pi,\pi]$. We learn 3 different fully connected NNDMs to show the generalizability of our method, which are: I. 3-layer with 50 hidden neurons per layer. II. 3-layer with 100 hidden neurons per layer. III. 4-layer with 50 hidden neurons per layer. Scalability analysis with more models can be found in \cref{apd:scale}.

When evaluating the control performance, we roll-out the closed-loop trajectory directly using NNDM to avoid model mismatch.
Our evaluation aims to show that the proposed method can provide provably safe controls efficiently for the learned model. The safe control computed by NNDM may be unsafe for the actual dynamics under model mismatch. We will extend our work to robust safe control~\cite{liu2015safe, noren2019safe}, which can guarantee safety even with model mismatch in the future.


To answer the questions we raised in the beginning, we design the following two tasks. The first task is trajectory tracking without safety constraints, which can test how our method performs comparing to other methods in terms of optimality and computational efficiency. And the second task is trajectory tracking under safety constraints. It is to test whether the learned safety index improves the feasibility and whether MIND-SIS ensures forward invariance and finite-time convergence.

\begin{table}[ht]
    \centering
    \scriptsize
    \begin{tabular}{c c c c c c c c c c}
    \toprule
        &   \multicolumn{3}{c}{NNDM I}   &   \multicolumn{3}{c}{NNDM II}   &   \multicolumn{3}{c}{NNDM III}\\
        \cmidrule(r){2-4} \cmidrule(r){5-7} \cmidrule(r){8-10}
        Method & Mean & Std & Time (s) & Mean & Std & Time (s) & Mean & Std & Time (s)\\
        \midrule
        MIND &$\mathbf{<10^{-8}}$ & $\mathbf{<10^{-7}}$ & 0.364 & $\mathbf{<10^{-8}}$ & $\mathbf{<10^{-7}}$ & 0.838 & $\mathbf{<10^{-8}}$ & $\mathbf{<10^{-7}}$ & 1.235\\
        Shooting-$10^3$ & 0.129 & 0.080 & 0.021 & 0.128 & 0.080 & 0.100 & 0.128 & 0.080 & 0.029\\
        Shooting-$10^4$ & 0.041 & 0.026 & 0.209 & 0.041 & 0.026 & 1.002 & 0.041 & 0.026 & 0.283\\
        Shooting-$10^5$ & 0.012 & 0.007 & 2.084 & 0.012 & 0.007 & 10.063 & 0.012 & 0.007 & 2.822\\
        Ipopt & 1.871 & 0.626 & 0.032 & 1.852 & 0.619 & 0.040 & 1.865 & 0.623 & 0.033\\
    \bottomrule
    \end{tabular}
    \caption{Average tracking error and average computation time of different methods in the trajectory tracking task (without safety constraint). The table shows mean and standard deviation of the average tracking error. The number after ``Shooting" denotes the sampling size. 
    Our method can always find the optimal solution, therefore achieves almost zero tracking error. The actual trajectories are illustrated in \cref{fig:tracking}.}
    \vspace{-5mm}
    \label{tab:tracking}
\end{table}

\subsection{Trajectory tracking}

In this task, we randomly generate 500 reference trajectory waypoints for each NNDM by rolling out the NNDM with some random control inputs. We compare our method (MIND) with 1) shooting methods with different sampling sizes and 2) Interior Point OPTimizer (Ipopt), a popular nonlinear solver. We use CPLEX to solve the MIND formulation. This experiment is done on a computer with AMD® Ryzen threadripper 3960x 24-core processor, 128 GB memory. Some results are shown in \cref{fig:tracking}.  Detailed settings and comparison of control sequences can be found in \cref{apd:tracking_setting}.

As shown in \cref{tab:tracking}, MIND achieves an average tracking error less than $10^{-8}$. The tracking error is less than the resolution of single-precision floats. We can conclude that MIND finds the optimal solution, which is a significant improvement comparing to other methods. The shooting method achieves lower tracking error with a larger sampling size but that also takes longer. To achieve the same tracking error as MIND, the sampling size and computation time will be unacceptably large. Ipopt takes shorter time because it gets stuck at local optima quickly. 

\begin{figure}[t]
     \centering
     \subfigure{\includegraphics[width=45mm]{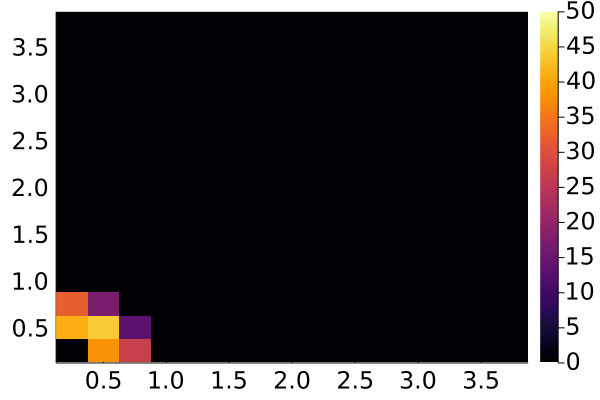}}
     \hfill
     \subfigure{\includegraphics[width=45mm]{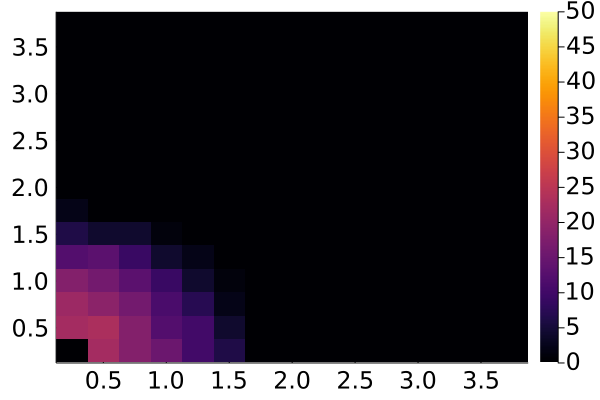}}
     \hfill
     \subfigure{\includegraphics[width=45mm]{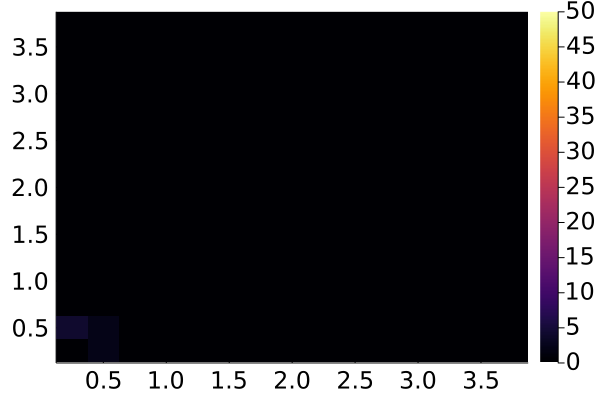}}
     \caption{Distribution of infeasible-state-of-interest states $B^*$ when we optimize the safety index on the whole state space ($B=X$). An obstacle is located at $(0,0)$. Each grid in the graph corresponds to a location. We sample 100 states at each location (with different heading angle and velocity). The color denote how many states at this location are in $B^*$. (a) shows that the original safety index $\phi_0$ has many infeasible states near the obstacle. (b) shows that a handcrafted safety index $\phi^h$ can not find feasible control for some states near the obstacle. (c) shows that the learned index $\phi$ has no state in $B^*$, thus can always find a feasible safe control.}
    \label{fig:safety_index}
\end{figure}


\subsection{Trajectory tracking under safety constraints}

This task considers two safety constraints corresponding to different scenarios: collision avoidance and safe following. When control infeasibility happens for a poorly designed safety index, we relax the safety constraint by adding a slack variable.

\subsubsection{Collision avoidance}

Collision avoidance is one of the most common safety requirement in real-world applications. In this experiment, we consider one static obstacle. The safety index is given as $\phi_0(\v x) = d_{min} - d(\v x) < 0$, where $d(\v x)$ is the relative distance from the agent to the obstacle, and $d_{min}$ is a constant. 
This constraint usually can not guarantee persistent feasibility. Therefore, we synthesize a safety index $\phi(\v x)$ that guarantees persistent feasibility by learning parameters of the following form:
$$\phi(\v x) = d_{min}^{\alpha_1} - d(\v x)^{\alpha_1} - \alpha_2\dot d(\v x) + \beta,$$
where $\dot d(\v x)$ is the relative velocity, $\alpha_1$, $\alpha_2$ and $\beta$ are parameters to learn. This form guarantees forward invariance and finite-time convergence for second-order systems when there is no control limits as shown in~\cite{liu2014control}, (see \cref{apd:Xs} for more discussion). The learned index can generalize to multiple obstacles case when we consider one constraint to each obstacle.

The search ranges for the parameters are: $\alpha_1 \in (0.1, 5)$, $\alpha_2 \in (0.1, 5)$, $\beta \in (0.001, 1)$.
For each set of parameters, we test how many sampled states are in $B^*$. We place an obstacle at $(0,0)$, and uniformly sample 40000 states around the obstacle to find $S$, then test whether the safe control set of each state is empty. \Cref{fig:safety_index} shows the distribution of $B^*$ of $\phi_0$, a manually tuned safety index $\phi^h$ given by previous work~\cite{liu2014control} ($\alpha_1 = 2, \alpha_2 = 1, \beta=0.1$, did not consider control limits), and a synthesized safety index $\phi^l$ with learned parameters ($\alpha_1 = 0.172, \alpha_2 = 4.107, \beta=0.447$). $\phi^l$ achieves $0$ infeasible rate. Additional comparison is in \cref{apd:Xs}. To demonstrate the effect of the synthesized safety index, we visualize the behavior of the agent with $\phi_0$ and $\phi^l$ in \cref{fig:obstacle}. The figure also shows that the synthesized safety index can be directly applied to unseen multi-obstacle scenarios without any change. The persistent feasibility is preserved if there is always at most one obstacle becoming safety critical~\cite{zhao2022model}. 

We evaluate these safety indices on 100 randomly generated collision avoidance tasks. The agent has to track a trajectory while avoiding collision (keep $\phi_0 \leq 0$). A task succeeds if there is no collision and no control-infeasible states throughout the trajectory. The evaluation results are shown in \cref{tab:feasibility}. The learned safety index achieves $0\%$ $\phi_0$-violation rate and $0\%$ infeasible rate. It is worth mentioning that \cite{zhao2022model} proposes a safety index design rule that guarantees feasibility for 2D collision avoidance. We verified that the $\phi^l$ satisfies this rule. 


\begin{table}[ht]
    \centering
    \footnotesize
    \begin{tabular}{c c c c c c c}
    \toprule
        &   \multicolumn{3}{c}{Collision avoidance} &   \multicolumn{3}{c}{Safe following}\\
        \cmidrule(r){2-4} \cmidrule(r){5-7}
        Metric & $\phi_0$ & $\phi^h$ & $\phi^l$ & $\phi_0$ & $\phi^h$ & $\phi^l$\\
        \midrule
        Success rate & $0\%$ & $89\%$ & $\mathbf{100\%}$ & $43\%$ & $82\%$ & $\mathbf{100\%}$\\
        $\phi_0$-violation rate & $100\%$ & $0\%$ & $0\%$ & $56\%$ & $0\%$ & $0\%$\\
        Infeasible rate & $100\%$ & $11\%$ & $0\%$ & $57\%$ & $18\%$ & $0\%$\\
    \bottomrule
    \end{tabular}
    \caption{Performance comparison of the original safety index $\phi_0$, a manually tuned safety index $\phi^h$, and a learned safety index $\phi^l$ on 100 randomly generated tasks. One trial is successful if there is no safety violation or infeasible state. We can see that $\phi_0$ always violates the constraints due to infeasibility (due to our choice of the initial state). $\phi^h$ fails to find control for some states due to infeasibility. Finally, the learned index $\phi^l$ can always find a safe control and avoid $\phi_0$ violation.}
    \vspace{-5mm}
    \label{tab:feasibility}
\end{table}

\subsubsection{Safe following}

The safe following constraint appears when an agent is following a target while keeping a safe distance, such as in adaptive cruise control, nap-of-the-earth flying, etc. The initial safety index is $\phi_0(\v x) = (d(\v x) - d_{l})(d(\v x) - d_{u})$, where $d_{l}$ and $d_{u}$ are the lower and upper bound of the relative distance.  We design the safety index to be of the form:
\begin{align}
    \phi(\v x) = \left|d(\v x) - \frac{ d_{l}+ d_{u}}{2}\right|^{\alpha_1} - \left(\frac{d_{u} - d_{l}}{2} + \beta\right)^{\alpha_1} + \alpha_2 \left[2d(\v x) - (d_{l} + d_{u})\right] \dot d(\v x).
\end{align}
The search range for the parameters to learn are $\alpha_1 \in (0.1,10), \alpha_2 \in (0.1,10)$, $\beta \in (0.001, 0.5)$. The learned parameters for $\phi^l$ are $\alpha_1 = 8.092, \alpha_2 = 9.826, \beta = 0.489$. The human designed parameters are $\alpha_1 = 2, \alpha_2 = 1, \beta = 0.01$ We also randomly generate 100 following tasks for evaluation. We consider a task successful if there is no $\phi_0$ constraint violation or infeasible state during the following. The evaluation results are shown in \cref{tab:feasibility}. The learned index achieves $0\%$ $\phi_0$-violation rate and $0\%$ infeasible rate.

\section{Discussion} \label{sec:discussion}

In this work, we propose MIND-SIS, the first method to derive safe control law for NNDM. MIND finds the optimal solution for safe tracking problems involving NNDM constraints, and SIS synthesizes a safety index that guarantees forward invariance and finite-time convergence. Theoretical guarantees of optimality and feasibility are provided. 
However, safety violation may still exist if the NNDM does not align with the true dynamics. As a future work, we will explore how to guarantee safety under model mismatch and uncertainty. One limitation of SIS is that to guarantee persistent feasibility, the theoretical sampling rate grows exponentially with the dimension of states. We will study how to adaptively adjust the sampling rate to overcome the curse of dimensionality.

\bibliography{reference}

\makeatletter
\@jmlrenddoc

\newpage
\appendix

\section{Appendices}

\subsection{Continuous time vs discrete time safe control} \label{apd:continuous_vs_discrete}
Our work is formulated in discrete time. There are vast literature dealing with safe control in continuous time. The major difference lies in the formulation of constraints. While our constraint is formulated as in \eqref{eq:safety}, the constraint for continuous time safe control is formulated as $\dot \phi\leq -\kappa(\phi)$ where $\kappa$ is a non decreasing function where $\kappa(0)\geq 0$. In CBF, $\kappa$ is chosen as $\kappa = \lambda\phi$. In SSA, $\kappa$ is chosen as $\gamma$ when $\phi\geq 0$ and $-\infty$ when $\phi<0$. Our method can be easily extend to continuous time safe control since $\dot\phi(\v x_k) \approx (\phi(\v x_{k+1})-\phi(\v x_k))/dt$. 

Due to the adoption of discrete time control, we approximate the safety constraints with first order Taylor expansion in \eqref{eq:taylor}. We define the error caused by first order approximation and omitting higher order terms of $\dot \phi$ as $e:=\phi(\v x_{k+1}) - \phi(\v x_k) - \nabla\phi(\v x_k) \v f(\v x_k, \v u_k) dt$.

\begin{lemma}\label{lemma: error bound}
The error $e$ can be bounded by  a constant safety margin $\varepsilon = 2 k_{\phi} M_f dt$ where $k_\phi$ is the Lipschitz constant of $\phi$, $M_f$ is the upper bound of $\|\v f(\v x, \v u)\|$ defined in Assumption $1$.
\end{lemma} 
\begin{proof}
\begin{subequations}
\begin{align}
    |e| 
    &= |\nabla\phi(\v x_k) \v f(\v x_k, \v u_k) dt - [\phi(\v x_{k+1}) - \phi(\v x_k)]|\\
    &\leq |\nabla\phi(\v x_k) \v f(\v x_k, \v u_k) dt| + k_\phi \|\v x_{k+1} - \v x_k\|\\
    & \leq \|\nabla\phi(\v x_k) \| \|\v f(\v x_k, \v u_k)\| dt + k_\phi \|\v f(\v x_k, \v u_k) dt\|\\
    & \leq  k_{\phi} M_f dt + k_{\phi} M_f dt = 2 k_{\phi} M_f dt = \varepsilon.
\end{align}
\end{subequations}
\end{proof}

Next, we show how the forward invariance and finite time convergence to the set $\{\v x:\phi^*(\v x)\leq 0\}$ defined by a safety index $\phi^*$ can be preserved in a discrete time system with the first order approximation of the safety constraint, by introducing a more conservative safety index that considers the approximation error. Note that one way to ensure forward invariance and finite time convergence to the set $\{\v x:\phi^*(\v x)\leq 0\}$ in discrete time is to find a control that satisfies
\begin{align}\label{eq: original control}
    \phi^*(\v x_{k+1}) \leq \max\{0, \phi^*(\v x_{k}) - \gamma dt\}.
\end{align}
However, in \eqref{eq: mixed integer program}, since we use the first order approximation of the constraint
\begin{align}
    \nabla_{\v x}\phi^* \cdot \v f(\v x_k, \v u_k) \leq  \max\{-\dfrac{\phi^*(\v x_k)}{dt},-\gamma\}, \label{eq:phi*_first}
\end{align}
the forward invariance and finite time convergence may not be preserved due to the approximation error. That is, a $\v u_k$ that satisfies \cref{eq:phi*_first} may lead to a $\phi^*(\v x_{k+1}) > \max\{0, \phi^*(\v x_{k}) - \gamma dt\}$. Therefore, to preserve forward invariance and finite time convergence to the set $\{\v x:\phi^*(\v x)\leq 0\}$, we introduce a new safety index (which is more conservative)
\begin{align}
    \phi:=\phi^*+\beta,
\end{align} and find the control based on \cref{eq:phi_first}:
\begin{align}
    \nabla_{\v x}\phi \cdot \v f(\v x_k, \v u_k) \leq  \max\{-\dfrac{\phi(\v x_k)}{dt},-\eta\}. \label{eq:phi_first}
\end{align}
The following lemma proves that this method ensures the forward invariance and finite time convergence to the set $\{\v x:\phi^*(\v x)\leq 0\}$.

\begin{lemma}\label{lemma: perservation}
If $\beta\geq\varepsilon$ and $\eta \geq \gamma + \varepsilon / dt$. Then a safe control $\v u_k$ that satisfies \eqref{eq:phi_first} also satisfies \eqref{eq: original control}.  
\end{lemma}

\begin{proof}
According to \lemmaref{lemma: error bound}, after applying control in \eqref{eq:phi_first}, we have $\phi(\v x_{k+1}) \leq \max\{0, \phi(\v x_{k}) - \eta dt\} + \varepsilon$, which implies that
\begin{align}
    \phi(\v x_{k+1}) - \varepsilon &\leq \max\{0,  \phi(\v x_{k})   - \eta dt\}.
\end{align}
By replacing $\phi$ with $\phi^* + \beta$, we further have $\phi^*(\v x_{k+1}) + \beta - \varepsilon \leq \max\{0,  \phi^*(\v x_{k}) + \beta - \eta dt\}$. Hence
\begin{align}
    \phi^*(\v x_{k+1})  &\leq \max\{\varepsilon - \beta ,   \phi^*(\v x_{k}) +\varepsilon - \eta dt\}\label{eq:eps-beta}\\  &\leq \max\{0 ,   \phi^*(\v x_{k}) - \gamma dt\}.
\end{align}
\end{proof}

Note that \lemmaref{lemma: perservation} states that the control law \eqref{eq:phi_first} ensures forward invariance and finite time convergence to the set $\{\v x:\phi^*(\v x)\leq 0\}$. In the extreme case when we select $\beta=\varepsilon$, the control law \eqref{eq:phi_first} (which is derived using $\phi$) ensures forward invariance and finite time convergence to the set $\{\v x:\phi(\v x)-\varepsilon\leq 0\}$.

\subsection{Safety index design and properties}\label{apd:Xs}

In this section, we show that the control law \eqref{eq:phi_first} not only ensures forward invariance and finite time convergence to the set $\{\v x:\phi(\v x)-\epsilon\leq 0\}$, but also to a subset of the user-specified safe set $\mathcal{X}_0$. 
It is worth noting that according to our parameterization, the set $\{\v x:\phi(\v x)-\epsilon\leq 0\}$ includes states that $\phi_0>0$. 
Denote this 0-sublevel set of $\phi - \varepsilon$ by $\XX_\phi$. That means $\XX_\phi$ is not necessarily a subset of $\XX_0$. 
In the following discussion, we show that for a second order system, \eqref{eq:phi_first} will make the set $\XX_s := \XX_\phi \cap \XX_0$ forward invariant if we optimize the safety index around its zero-level set to the point that $B^*=\emptyset$; and the set will be finite-time convergent if we optimize the safety index for the whole state space (by letting the state-of-interest set $B=X$) to the point that $B^*=\emptyset$. 

\begin{figure}
    \centering
    \centering
     \subfigure{\includegraphics[width=.52\linewidth]{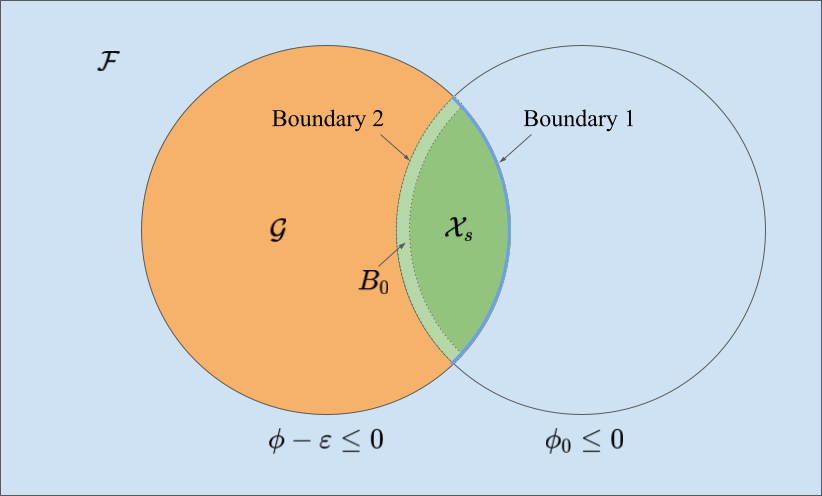} \label{fig:partition_a}}
     \hfill
     \subfigure{\includegraphics[width=.35\linewidth]{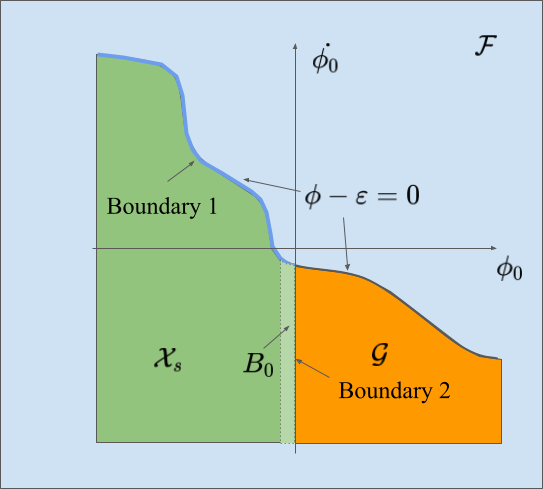} \label{fig:partition_b}}
    \caption{Space partition. The state space is partitioned into three parts. And the safe set is defined by two boundaries. (a) shows the partition in a Venn graph and (b) shows the partition in the phase graph.}
    \label{fig:apd2}
\end{figure}

To prove this, we first partition the whole state space into three parts as shown in \cref{fig:apd2}:
\begin{itemize}
    \item The space $\FF:= \{\v x\mid \phi(\v x) - \varepsilon>0\}$;
    \item The space $\XX_s:= \{\v x\mid \phi(\v x)  - \varepsilon\leq 0\}\cap \{\v x\mid \phi_0(\v x)\leq 0\}$;
    \item The space $\GG:=\{\v x\mid \phi(\v x)  - \varepsilon \leq 0\}\cap \{\v x\mid \phi_0(\v x)>0\}$.
\end{itemize}

We will prove the forward invariance and finite-time convergence to $\XX_s$ defined by the safety index we used in this work: $\phi(\v x) - \varepsilon = \phi_0^*(\v x) + \alpha\dot \phi_0(\v x) + \beta - \varepsilon$.

For forward invariance, it suffices to show that all $\Delta \v x = \v f(\v x, \v u)$ along the boundary of $\XX_s$ are pointing toward the interior of $\XX_s$. For finite-time convergence, we are going to show that under certain conditions, all trajectories starting from $\GG$ will converge to $\XX_s$ and all trajectories starting in $\FF$ will converge to $\XX_s\cup\GG$.

\subsubsection{Discrete time forward invariance of \texorpdfstring{$\XX_s$}{Lg}}

The boundary of $\XX_s$ can be decomposed into two parts $\{\v x \mid \phi(\v x) - \varepsilon = 0, \phi_0(\v x) \leq 0\}$ and $\{\v x \mid \phi(\v x) - \varepsilon \leq 0, \phi_0(\v x) = 0\}$.
In the above discussion, we proved that when $\phi(\v x_{k}) - \varepsilon < 0$, $\phi(\v x_{k+1}) - \varepsilon < 0$. Therefore the first part of boundary will not be crossed. We only need to consider the second part.
We first define the near-$\phi_0$-boundary-state set $B_0 = \{\v x \mid \phi(\v x) - \varepsilon \leq 0, \ \exists \v x_0,\ \phi_0(\v x_0) = 0, \text{ and } \|\v x - \v x_0\| < M_f dt \}$. Because $\|\v x_{k+1} - \v x_{k}\| \leq \|\v f(\v x, \v u)\| dt \leq M_f dt$. $B_0$ contains all the states that may lead to a out-of-boundary state in one step. Then we only need to prove that
\begin{align}
    \forall \v x_k \in B_0, \phi_0(\v x_{k+1}) < 0.
\end{align}

Similar to \cref{apd:continuous_vs_discrete}, we can derive the discretization error bound $\varepsilon_0 = 2 k_{\phi_0} M_f dt$ for $\phi_0$, and $\varepsilon_0^* = 2 k_{\phi_0^*} M_f dt$ for $\phi_0^*$ under the following assumption.
\begin{assumption}
$\phi_0$ and $\phi_0^*$ are both Lipschitz continuous with Lipschitz constants $k_{\phi_0}$ and $k_{\phi_0^*}$ respectively.
\end{assumption}

\begin{lemma}[Forward Invariance]\label{lemma: alpha}
If $\beta - \varepsilon - \varepsilon_0^*/2 > 0$, $\alpha < \frac{\beta - \varepsilon - \varepsilon_0^*/2}{\varepsilon_0} dt$, then $\forall \v x_k \in B_0, \phi_0(\v x_{k+1}) \leq 0$.
\end{lemma}
\begin{proof}
For $\v x_k\in B_0$, we have
\begin{align}
    \phi(\v x_k) - \varepsilon = \phi_0^*(\v x) + \alpha \dot \phi_0(\v x) + \beta - \varepsilon\leq 0,
\end{align}
and using Lipschitz continuity, we have that $\forall \v x \in B_0$, $-k_{\phi_0^*} M_f dt < \phi_0^*(\v x) < 0$ .  Therefore
\begin{align}
    \dot \phi_0(\v x) \leq \frac{-\phi_0^*(\v x) - \beta + \varepsilon}{\alpha} \leq \frac{ k_{\phi_0^*} M_f dt - \beta + \varepsilon}{\alpha} = \frac{ \varepsilon_0^*/2 - \beta + \varepsilon}{\alpha}.
\end{align}
Then based on Lemma 2, the following inequality holds for arbitrary $u_k$.
\begin{align}
    \phi_0(\v x_{k+1}) &\leq \phi_0(\v x_{k}) + \dot \phi_0(\v x_{k}) dt + \varepsilon_0.\\
\end{align} 
And because $\phi_0(\v x_k) \leq 0$ and $\alpha < \frac{\beta - \varepsilon - \varepsilon_0^*/2}{\varepsilon_0} dt$, we have
\begin{align}
    \phi_0(\v x_{k+1}) &\leq \phi_0(\v x_{k}) + \dot \phi_0(\v x_{k}) dt + \varepsilon_0\\
    &\leq \phi_0(\v x_{k}) + \frac{\varepsilon_0^*/2 - \beta + \varepsilon}{\alpha} dt + \varepsilon_0\\
    &\leq \frac{\varepsilon_0^*/2 - \beta + \varepsilon}{\alpha} dt + \varepsilon_0 \leq 0.
\end{align} \label{eq:phi_0_converge}
\end{proof}

\subsubsection{finite-time convergence of \texorpdfstring{$\XX_s$}{Lg}}
We prove the finite-time convergence of $\XX_s$ based on the following assumption:
\begin{assumption}
We optimize the safety index for the whole state space by letting the state-of-interest set $B=X$. And the infeasible-state-of-interest set $B^*=\emptyset$. Therefore, $\phi-\varepsilon$ has feasible solution for the safety constraint for arbitrary state.
\end{assumption}
It is obvious that all trajectories start in $\FF$ will converge to $\XX_s \cup \GG$ in finite steps because $\phi(\v x_{k+1}) - \varepsilon< \phi(\v x_{k}) - \varepsilon - \eta$, $\phi - \varepsilon$ will decrease to $0$ in finite-time. Then we consider the trajectories start in $\GG$.
\begin{align}
    \phi(\v x_{k})-\varepsilon = \phi_0^*(\v x_{k}) + \alpha \dot \phi_0(\v x_{k}) + \beta -\varepsilon \leq 0 \implies \dot \phi_0(\v x_{k}) \leq \frac{-\phi_0^*(\v x_{k}) -\beta + \varepsilon}{\alpha}.
\end{align}
Note that $\phi_0^*(\v x_{k}) > 0,\ \forall \v x \in \GG$ because $\phi_0^*$ defines the same sublevel set as $\phi_0$. Therefore
\begin{align}
    \dot \phi_0(\v x_{k}) \leq \frac{-\phi_0^*(\v x_{k}) -\beta+ \varepsilon}{\alpha} \leq \frac{-\beta+ \varepsilon}{\alpha}.
\end{align}
Then based on \lemmaref{lemma: error bound}, the following inequality holds for arbitrary $\v u_k$.
\begin{align}
    \phi_0(\v x_{k+1}) & \leq \phi_0(\v x_{k}) + \dot \phi_0(\v x_{k}) dt + \varepsilon_0\\
    &\leq \phi_0(\v x_{k}) + \frac{ - \beta + \varepsilon}{\alpha} dt + \varepsilon_0\\
    &\leq \frac{ - \beta + \varepsilon}{\alpha} dt + \varepsilon_0.
\end{align}
If we choose $\alpha$ according to \lemmaref{lemma: alpha}, then the righthand side is a negative constant.
$\phi_0$ decreases to $0$ in finite-time, therefore the trajectory will converge to $\XX_s$ in finite-time.

\subsubsection{Phase plots} \label{apd:phase}

\begin{figure}[t]
     \centering
     \subfigure[$\phi_0$ phase plot]{\includegraphics[width=40mm]{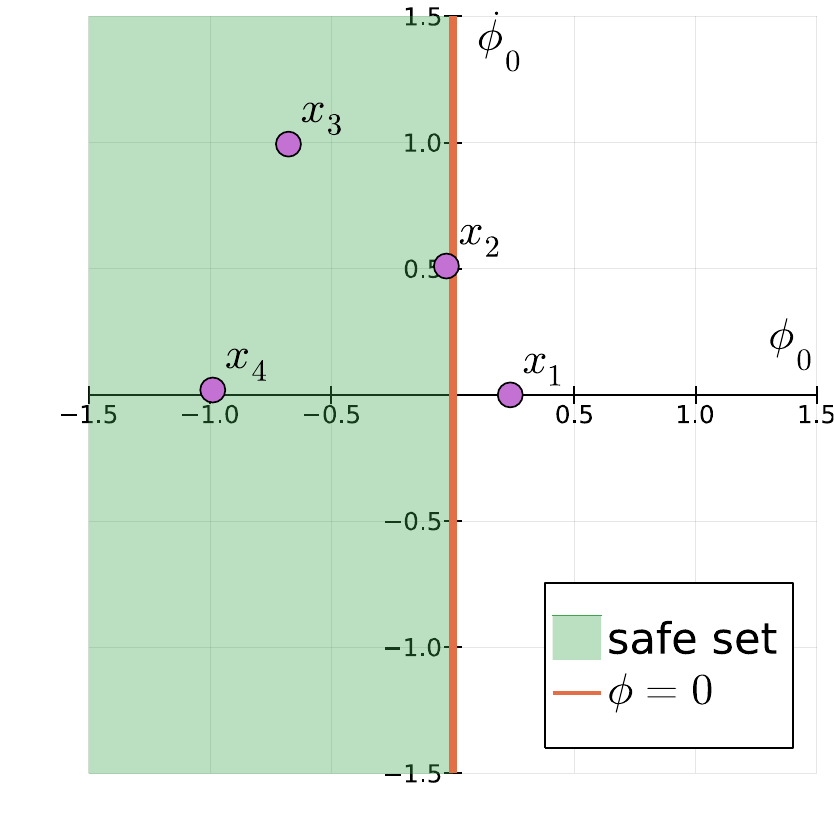}}
     \hfill
     \subfigure[$\phi^h$ phase plot]{\includegraphics[width=40mm]{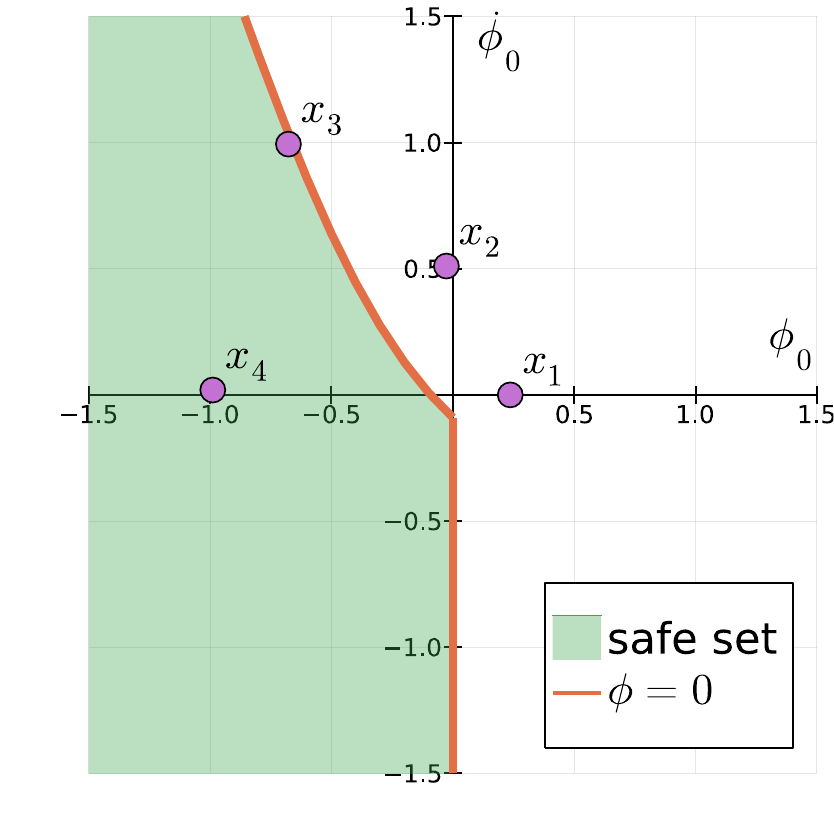}}
     \hfill
     \subfigure[$\phi^l$ phase plot]{\includegraphics[width=40mm]{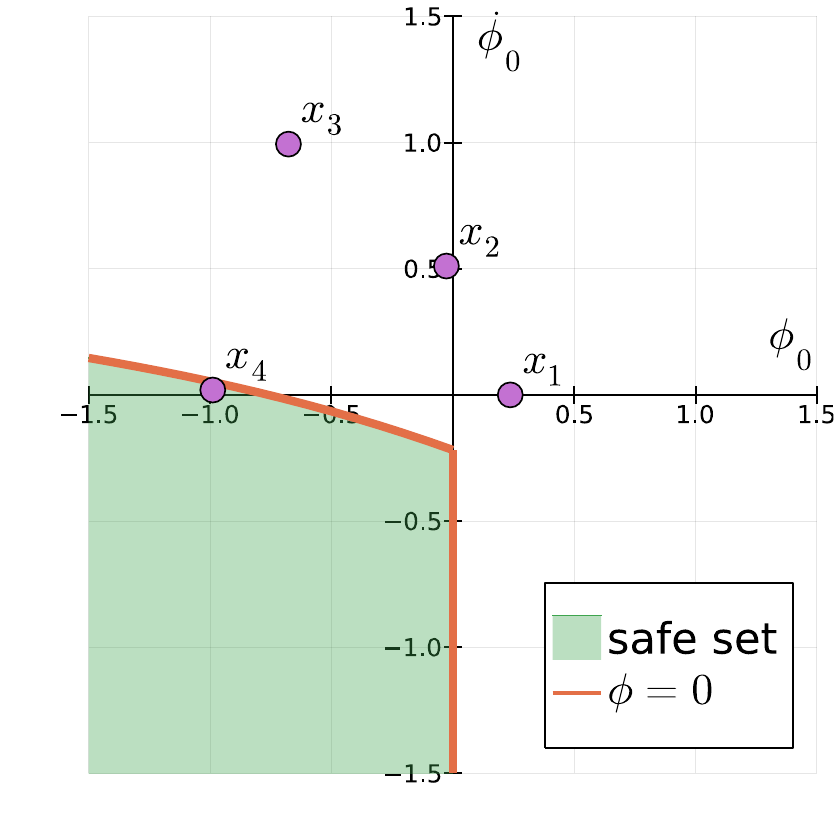}}
     \newline
     \subfigure[$\phi_0$ control feasibility]{\includegraphics[width=40mm]{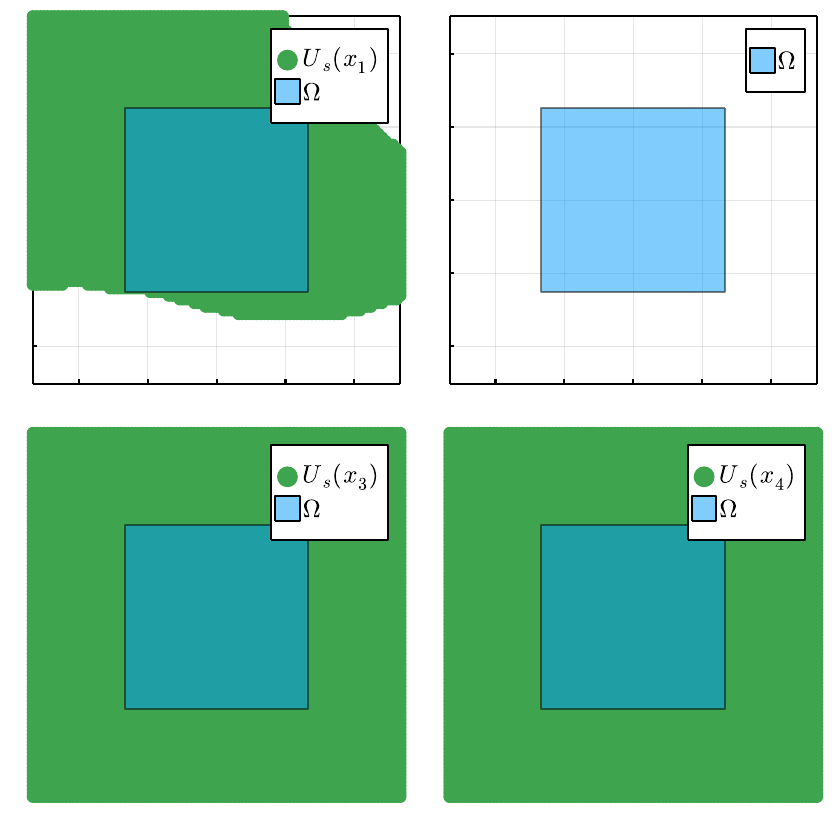}}
     \hfill
     \subfigure[$\phi^h$ control feasibility]{\includegraphics[width=40mm]{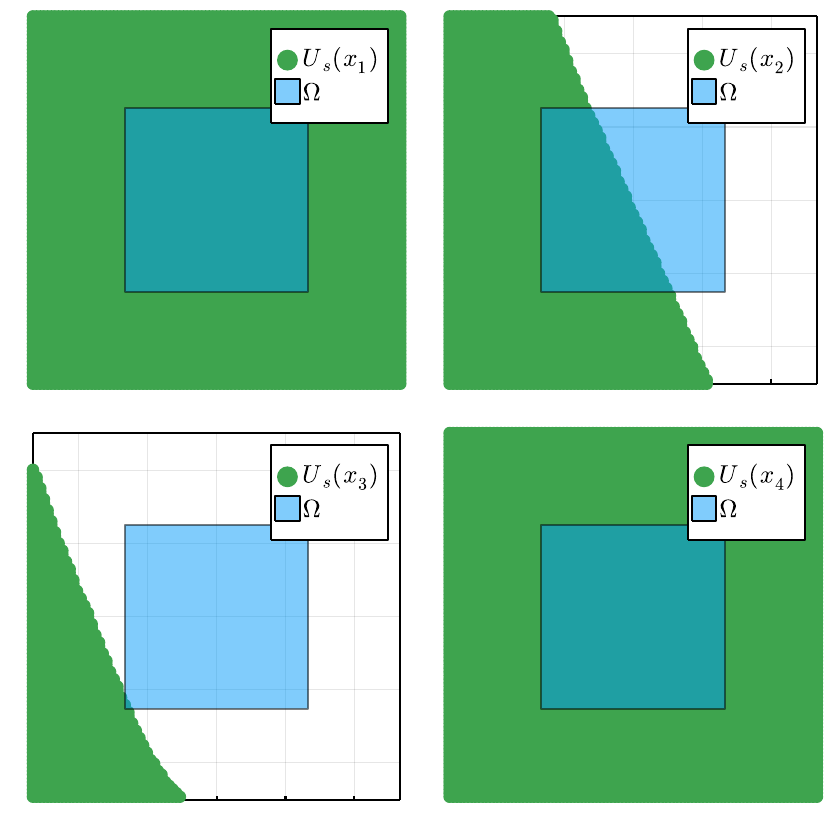}}
     \hfill
     \subfigure[$\phi^l$ control feasibility]{\includegraphics[width=40mm]{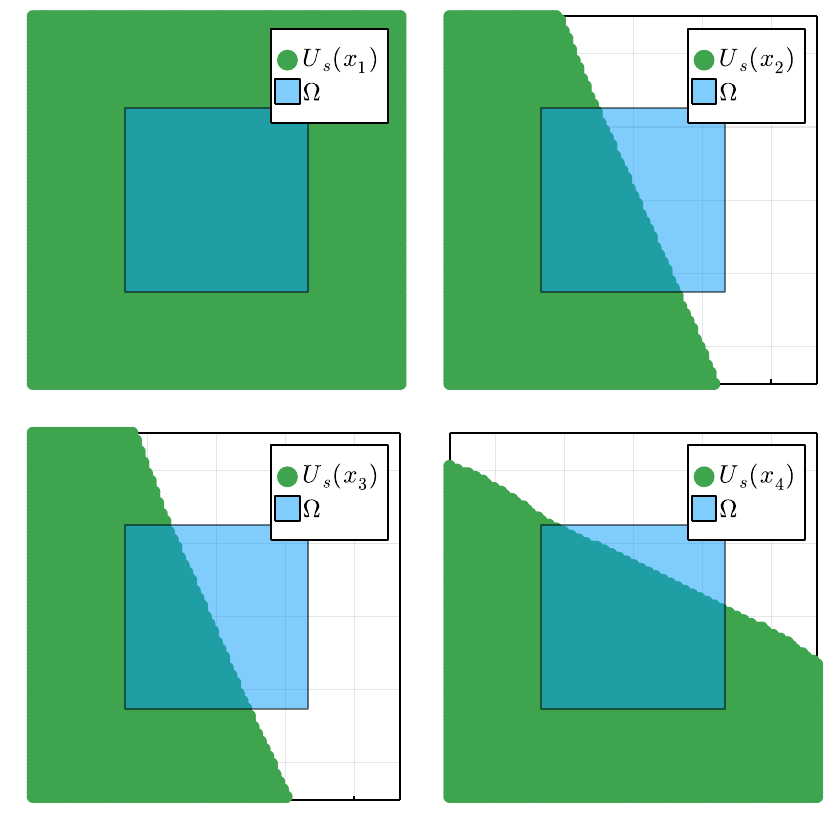}}
     \centering
    \caption{Phase plots and control feasibility plots. The first row shows the phase plot for different safety index. And the second row shows the control spaces at four sampled states $\v x_1$ to $\v x_4$. The blue squares denotes the control limit $\Omega$. The green areas are the feasible controls that satisfy the safety constraint. Before the safety index synthesis, there is no feasible control for $\v x_2$. And for the manually designed safety index $\phi^h$, there is almost no feasible control for $\v x_3$. But for the learned safety index $\phi^l$. The feasibility is guaranteed for arbitrary states.}
    \label{fig:phase}
\end{figure}

Phase plot shows the trajectories of the system dynamics in the phase plane. We can see how different safety indexes reacts to the same situation. We draw the trajectory of $\phi=0$ for $\phi_0$, $\phi^h$, and $\phi^l$, as shown in \cref{fig:phase}. $\phi^l$ is more conservative, characterizes a smaller safe set, but ensures feasibility of all states.

\subsection{Tracking with NNDM}\label{apd:tracking_setting}

\begin{figure}[t]
     \centering
     \subfigure{\includegraphics[width=40mm]{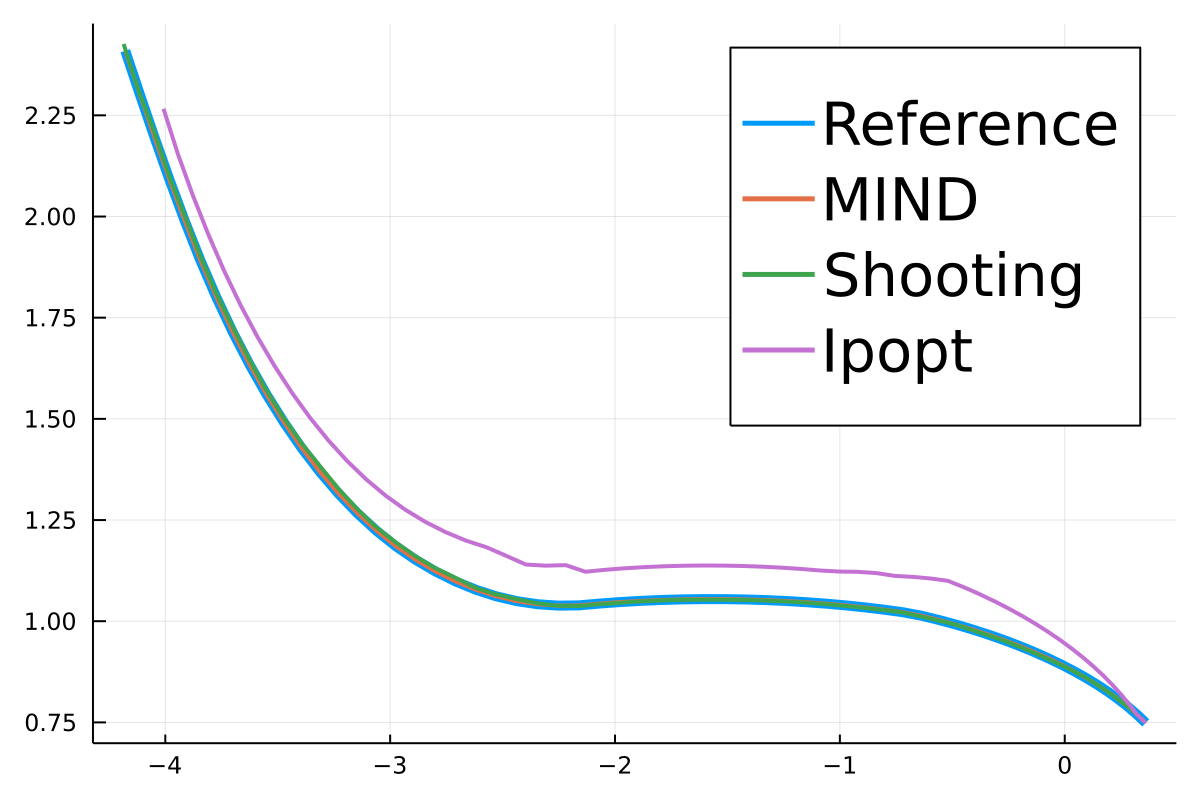}}
     \hfill
     \subfigure{\includegraphics[width=40mm]{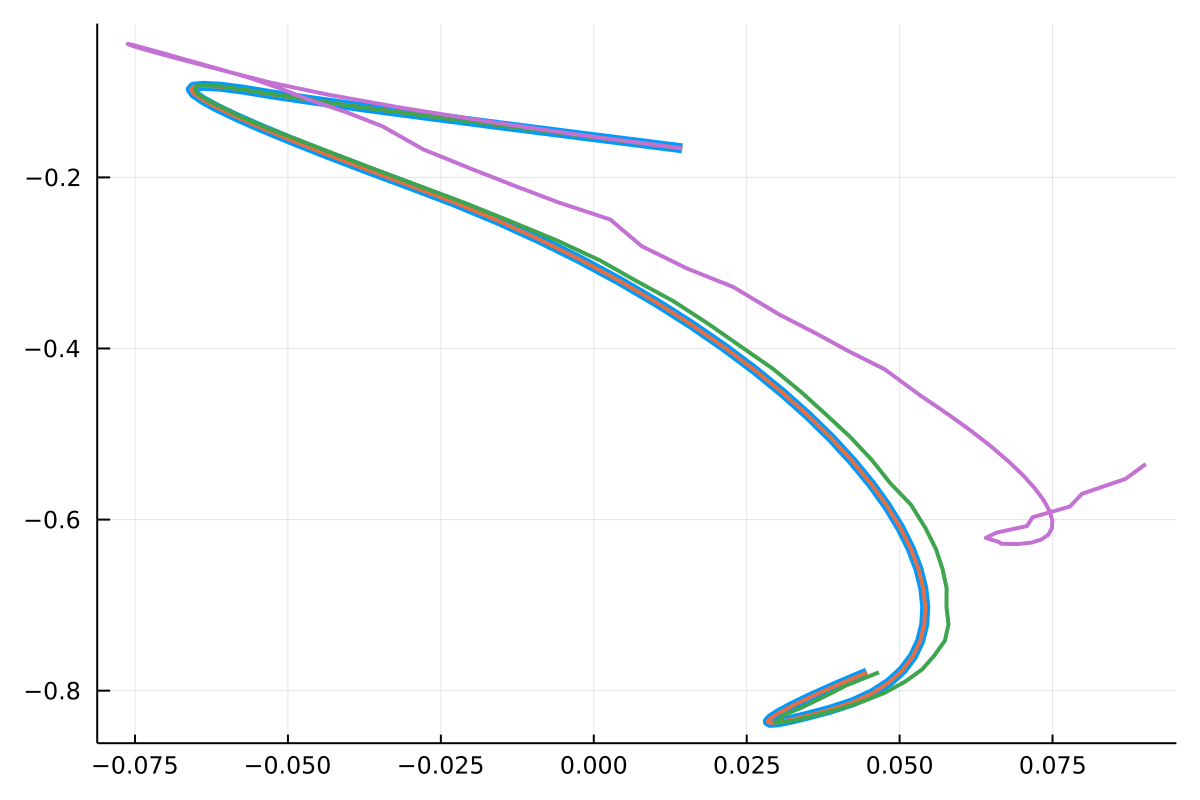}}
     \hfill
     \subfigure{\includegraphics[width=40mm]{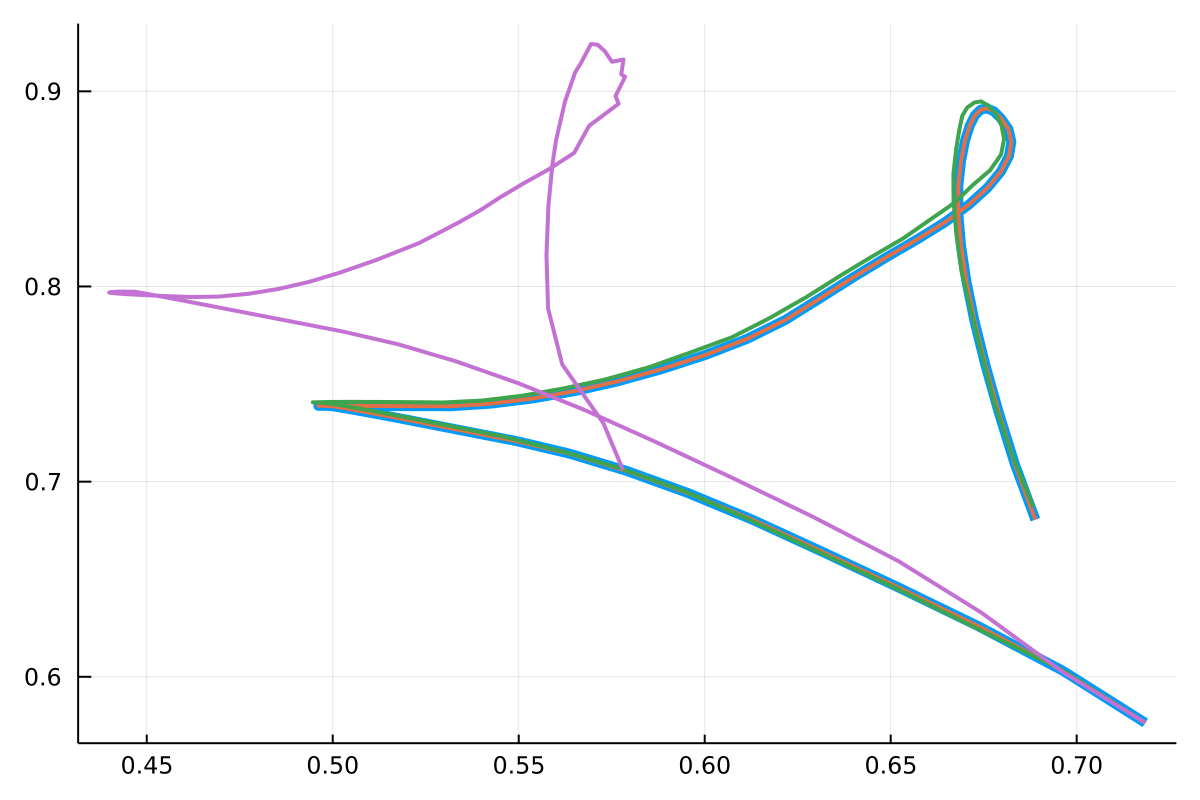}}
     \newline
     \subfigure{\includegraphics[width=40mm]{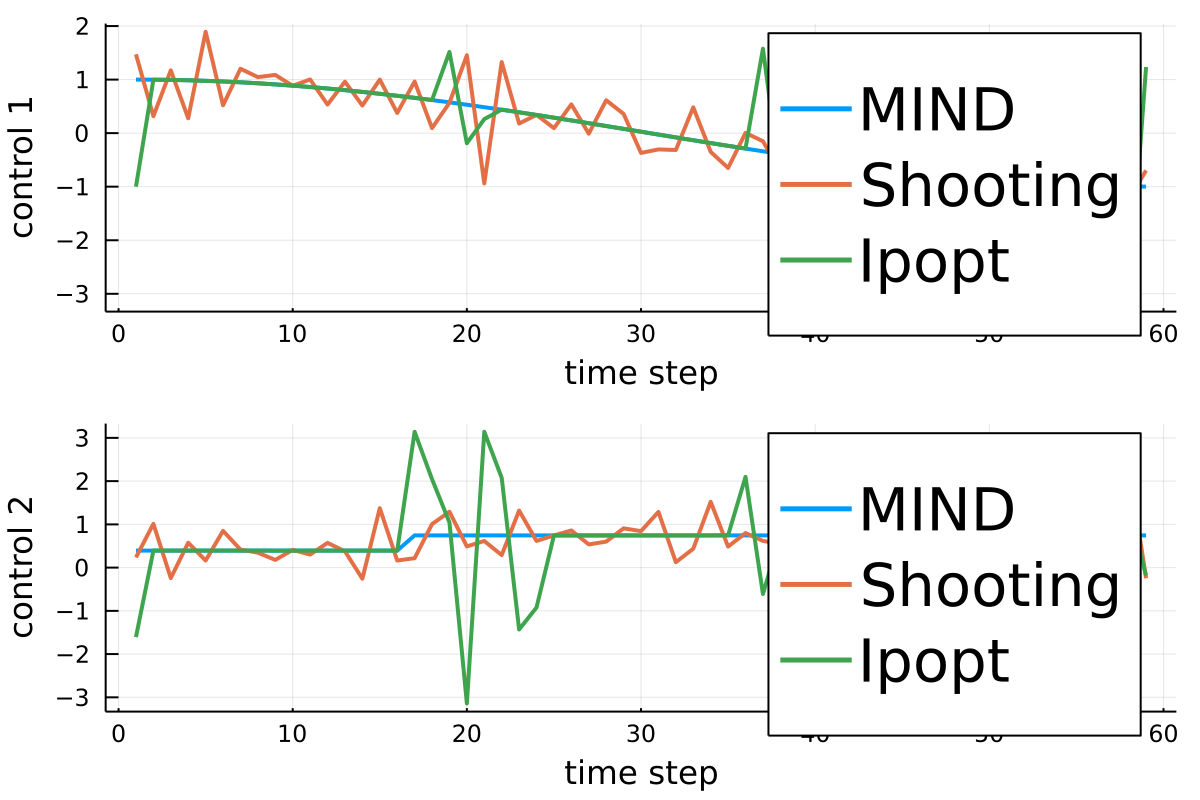}}
     \hfill
     \subfigure{\includegraphics[width=40mm]{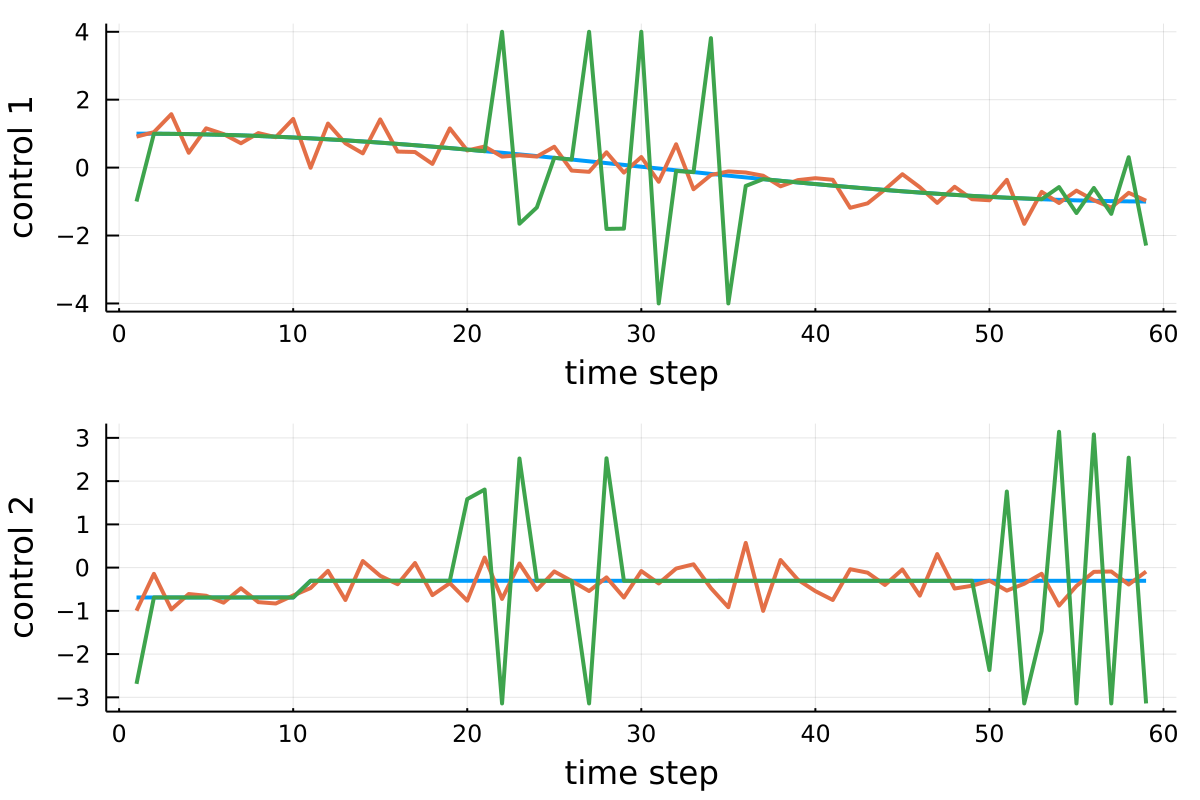}}
     \hfill
     \subfigure{\includegraphics[width=40mm]{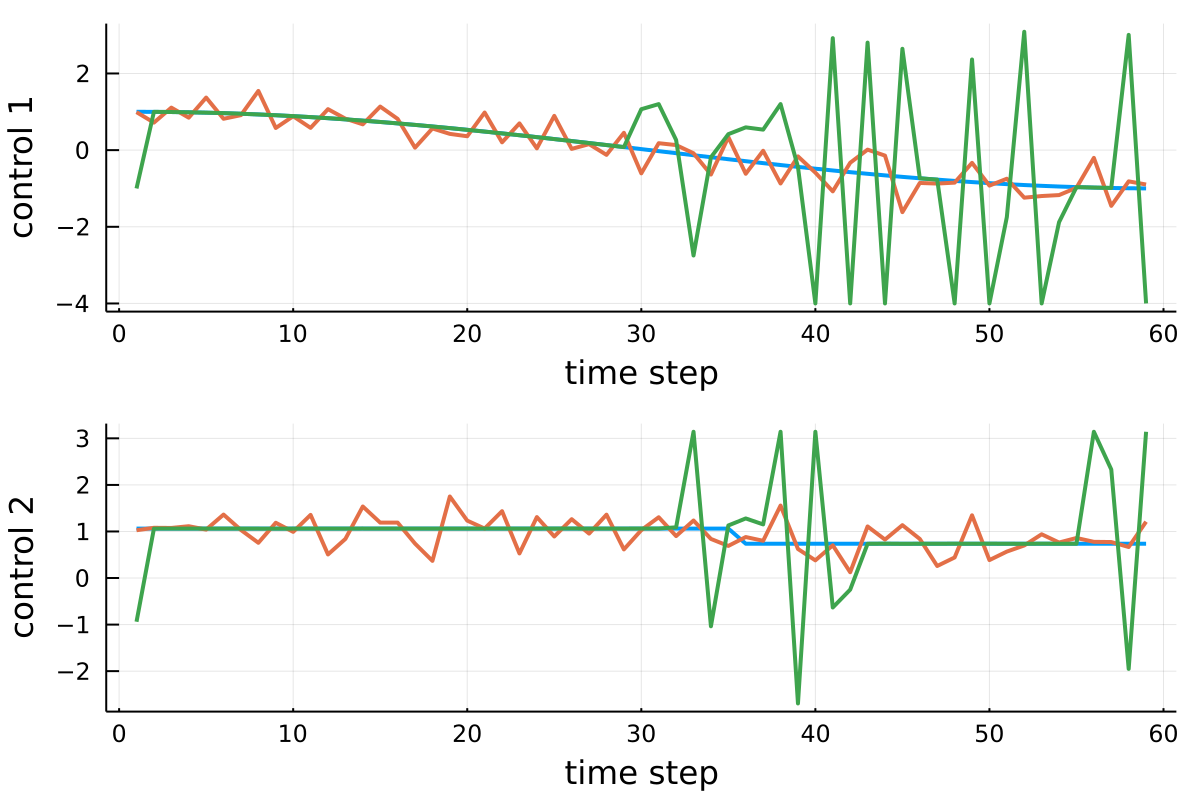}}
     \centering
    \caption{Trajectory tracking using MIND, shooting-100, and Ipopt. The first row shows the trajectories, and the second row shows two corresponding control signals for the trajectories.}
    \label{fig:jagged_traj}
\end{figure}

\begin{figure}[t]
     \centering
     \subfigure{\label{heat:a}\includegraphics[width=40mm]{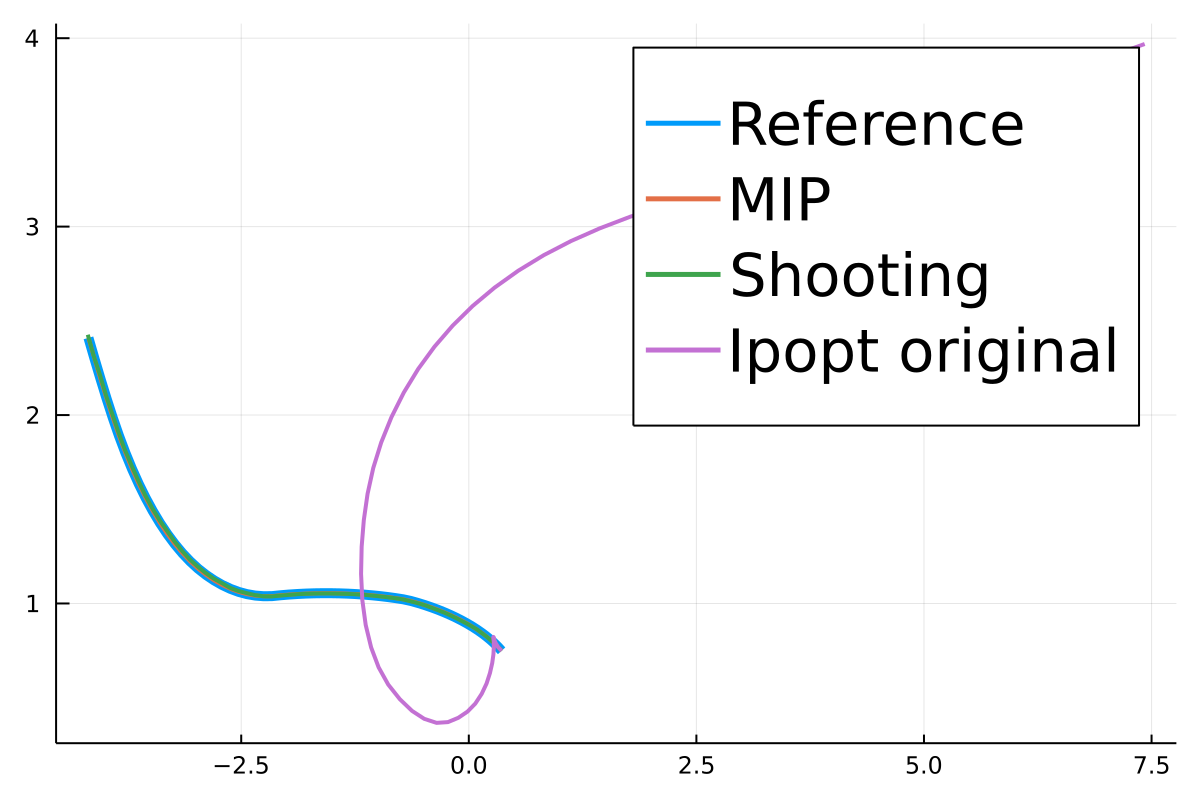}}
     \hfill
     \subfigure{\label{heat:b}\includegraphics[width=40mm]{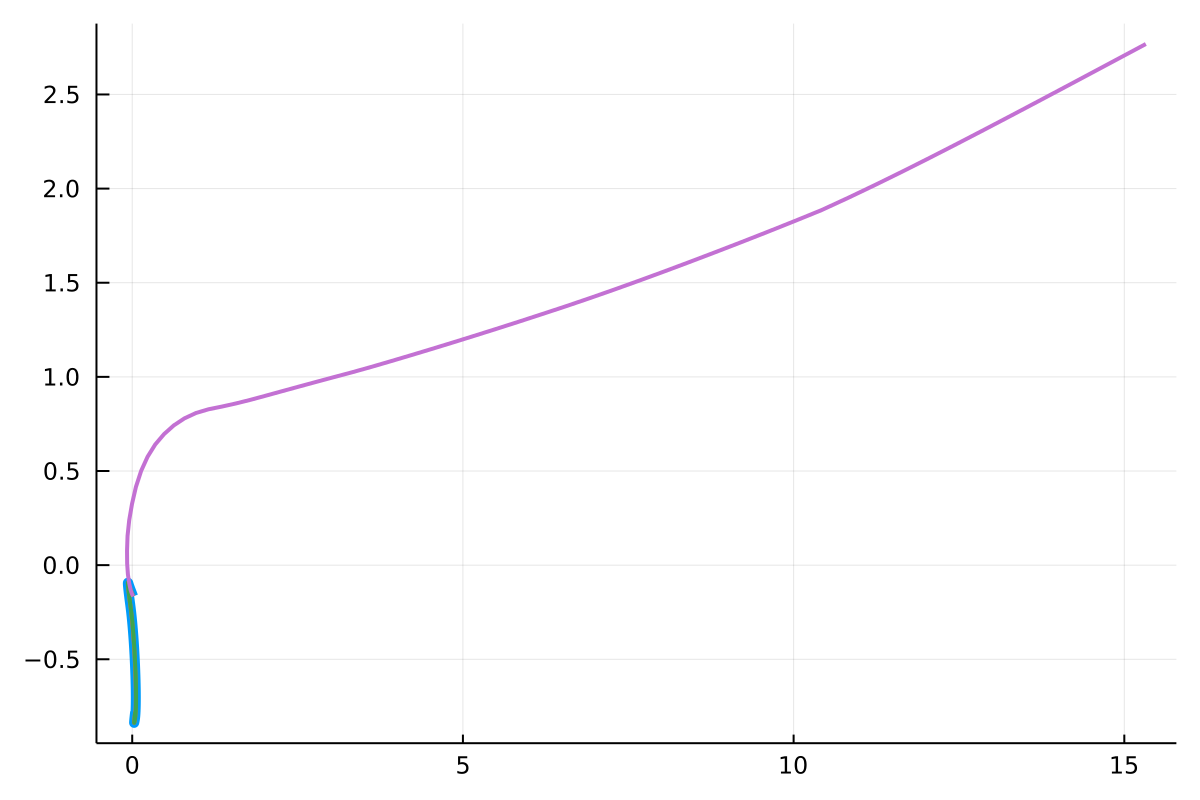}}
     \hfill
     \subfigure{\label{heat:c}\includegraphics[width=40mm]{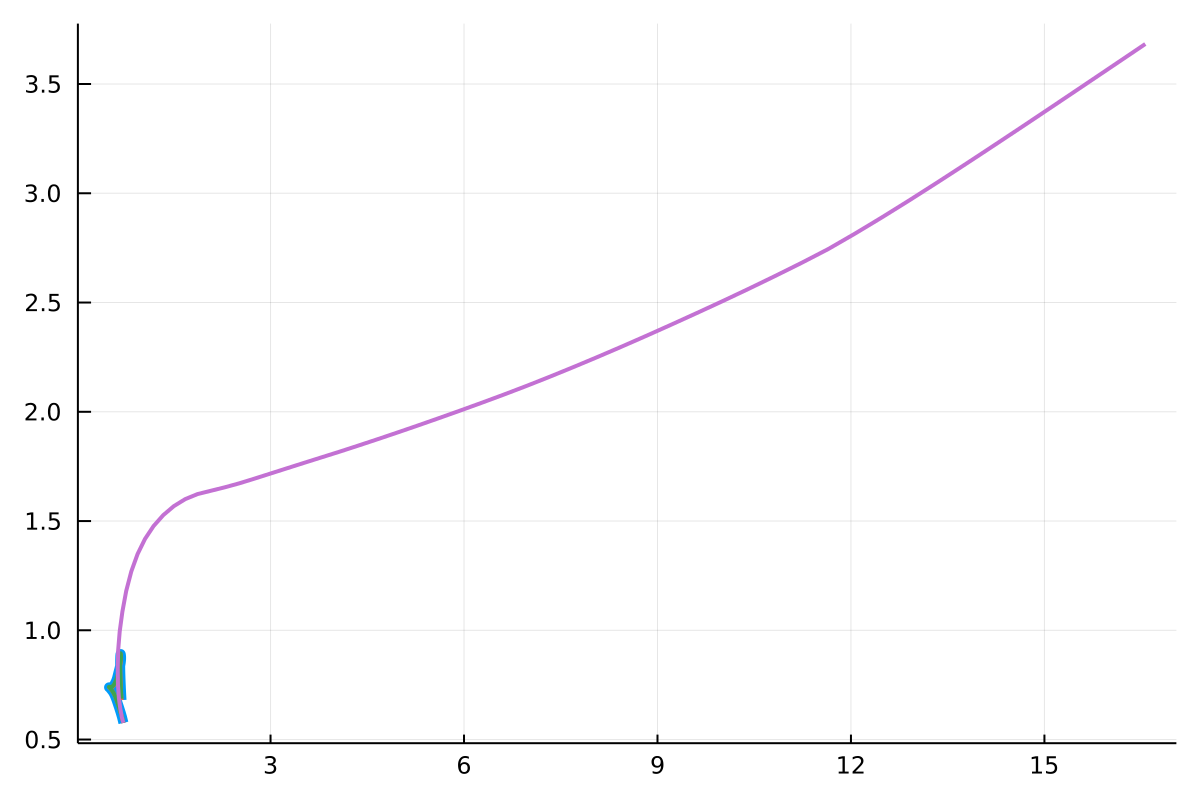}}
     \newline
     \subfigure{\label{heat:d}\includegraphics[width=40mm]{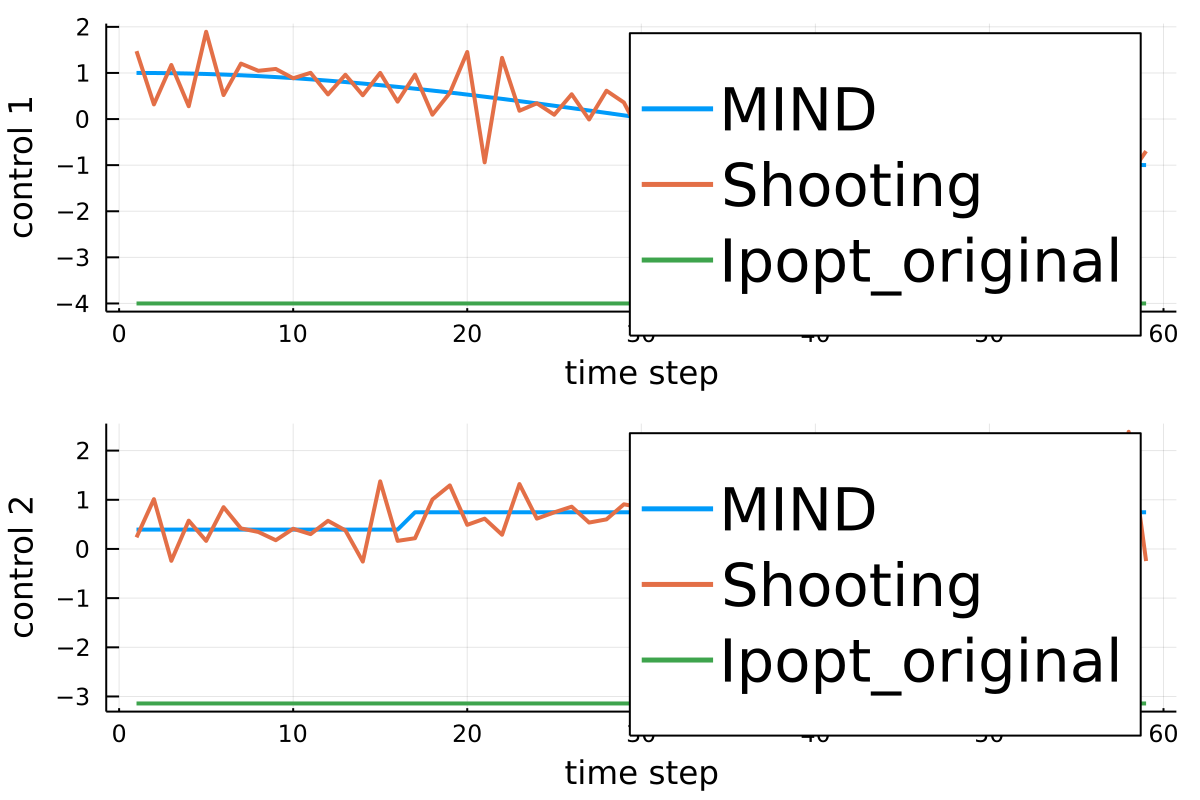}}
     \hfill
     \subfigure{\label{heat:e}\includegraphics[width=40mm]{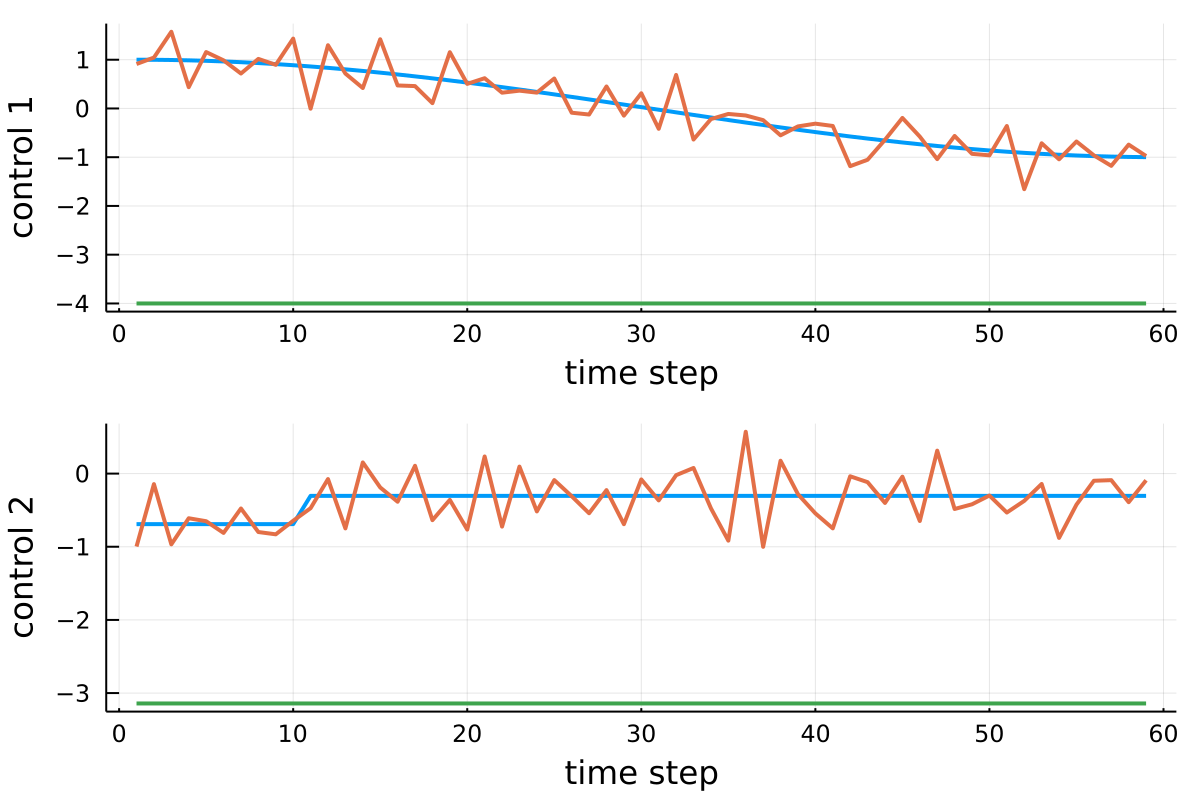}}
     \hfill
     \subfigure{\label{heat:f}\includegraphics[width=40mm]{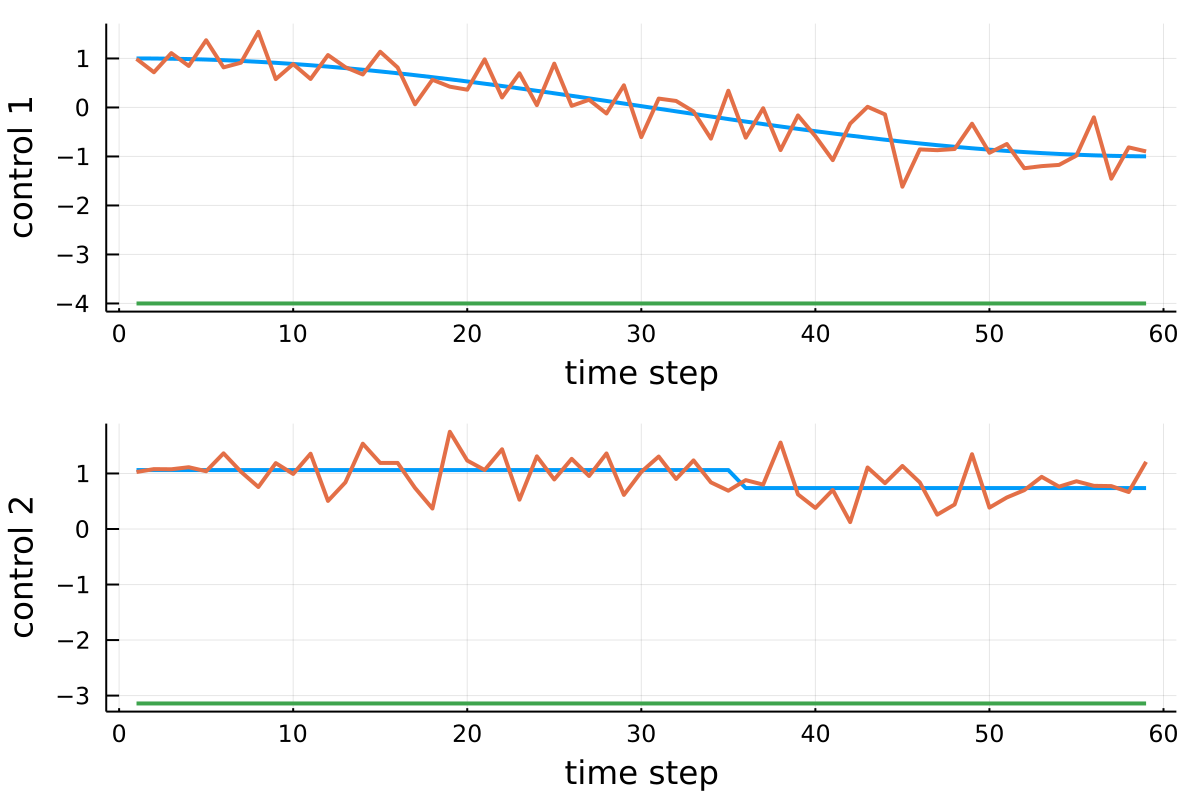}}
     \centering
    \caption{Trajectory tracking using MIND, shooting-100, and original Ipopt. The reference trajectories are the same as \cref{fig:jagged_traj}, But the original Ipopt trajectories deviate too much. The second row shows two corresponding control signals. Original Ipopt can not find feasible control.}
    \label{fig:original_ipopt}
\end{figure}

We compared tracking performance of different solvers. The performance of the original Ipopt solver is very inefficient, so we add an extra optimality constraint to help it find a better solution. Specifically, on top of \cref{eq:safe_control}, we add the following constraint:
\begin{align}
    |\v x_{k+1} - \v x_{k+1}^r| < \v x_{\epsilon},
\end{align}
where $\v x_{\epsilon}$ is a constant vector that we define to bound the optimized state. When the velocity term and orientation term of $\v x_{\epsilon}$ are small, Ipopt is able to find a smooth trajectory as shown in \cref{fig:tracking}. But if we use a larger bound for velocity and orientation, Ipopt can only find jagged trajectories because it often gets stuck at local optima. \cref{fig:jagged_traj} show the trajectories and two control signals when all terms of $\v x_e$ is $0.1$. 
The results of the original Ipopt without any additional constraints are shown in \cref{fig:original_ipopt}.

\subsection{MIND Scalability} \label{apd:scale}

We studied how MIND can scale to different model sizes. We test models with different layers and different hidden dims as shown in \cref{tab:scalability}. Besides the computation time, we also show the average prediction error of these models to demonstrate their relative learning ability. We may consider a model with a smaller error having a better learning ability. As a baseline, \cite{nagabandi2018neural} uses a 2-layer 500 hidden dim network to learn the model of an ant robot (one torso and four 3-DoF legs), which has 111 state dimensions, with joint torque as control input. And most models (whose error is less than 0.168) we test may have better learning ability than the 2-layer 500 hidden dim one. So it is reasonable to assume that these models are already enough to learn a variety of complex dynamics with high dimensional input output. It shows the potential of our method to achieve real time control for models with complex dynamics.



\begin{table}
    \centering\footnotesize
    \begin{tabular}{c c c c c c c}
    \toprule
        & \multicolumn{2}{c}{2-layer} & \multicolumn{2}{c}{3-layer} & \multicolumn{2}{c}{4-layer}\\
        \cmidrule(r){2-3} \cmidrule(r){4-5} \cmidrule(r){6-7}\\
        Hidden Dim & error & time & error & time & error & time\\
        \midrule
        50 & 0.275 & 0.033 s & 0.150 & 0.340 s & 0.141 & 0.884 s\\
        100 & 0.198 & 0.098 s & 0.138 & 1.654 s & 0.133 & Stopped\\
        200 & 0.188 & 0.294 s & 0.137 & Stopped& &\\
        300 & 0.190 & 0.696 s & & & &\\
        500 & 0.168 & Stopped & & & &\\
    \bottomrule
    \end{tabular}
    \caption{Average norm of prediction errors after 1000 epochs training on the vehicle dynamics, and average computation time of MIND for one step. We stopped the computation at $2$ s, beyond which point it can no longer be used for real-time control. 
    The prediction error norm is computed by $\|\dotv x - \dotv x_{Actual}\|$. We only use this error to show their relative learning ability. A smaller error may suggest the model having a better learning ability.}
    \label{tab:scalability}
\end{table}

\end{document}